\documentclass[Afour,sagev,times]{sagej}

\usepackage[dvipsnames]{xcolor}
\usepackage[normalem]{ulem}
\usepackage{moreverb}
\usepackage[hyphens]{url}
\usepackage{graphicx}
\usepackage{comment}
\DeclareGraphicsExtensions{.png,.jpg}
\usepackage{multirow}
\usepackage{multicol}
\usepackage{cancel}
\usepackage{flushend}

\usepackage[colorlinks,bookmarksopen,bookmarksnumbered,citecolor=red,urlcolor=red]{hyperref}

\newcommand\BibTeX{{\rmfamily B\kern-.05em \textsc{i\kern-.025em b}\kern-.08em
T\kern-.1667em\lower.7ex\hbox{E}\kern-.125emX}}

\def\volumeyear{2020}

\begin{document}

\runninghead{Clark \textit{et~al.}}

\title{Training Data Augmentation for Deep Learning Radio Frequency Systems}

\author{William H. Clark IV\affilnum{1}, Steven Hauser\affilnum{2}, William C. Headley\affilnum{1}, and Alan J. Michaels\affilnum{1}}

\affiliation{\affilnum{1}Ted and Karyn Hume Center for National Security and Technology, Blacksburg, VA\\
\affilnum{2}Adapdix Corporation, Pleasanton, CA}

\corrauth{William H Clark IV, Ted and Karyn Hume Center for National Security and Technology,
Virginia Tech,
Blacksburg, VA,
USA}

\email{saikou@vt.edu}

\begin{abstract}
Applications of machine learning are subject to three major components that contribute to the final performance metrics.
Within the category of neural networks, and deep learning specifically, the first two are the architecture for the model being trained and the training approach used.
This work focuses on the third component, the data used during training.
The primary questions that arise are ``what is in the data'' and ``what within the data matters?''
Looking into the Radio Frequency Machine Learning (RFML) field of Automatic Modulation Classification (AMC) as an example of a tool used for situational awareness, the use of synthetic, captured, and augmented data are examined and compared to provide insights about the quantity and quality of the available data necessary to achieve desired performance levels.
There are three questions discussed within this work: (1) how useful a synthetically trained system is expected to be when deployed without considering the environment within the synthesis, (2) how can augmentation be leveraged within the RFML domain, and lastly, (3) what impact knowledge of degradations to the signal caused by the transmission channel contributes to the performance of a system.
In general, the examined data types each have useful contributions to a final application, but captured data germane to the intended use case will always provide more significant information and enable the greatest performance.
Despite the benefit of captured data, the difficulties and costs that arise from live collection often make the quantity of data needed to achieve peak performance impractical.
This paper helps quantify the balance between real and synthetic data, offering concrete examples where training data is parametrically varied in size and source.
\end{abstract}


\maketitle

\section{Introduction}
Over the past decade, a significant amount of interest and investment have been poured into the development of Machine Learning (ML) algorithms and systems; this paper addresses the core areas of training data quantification and augmentation for Radio Frequency (RF) systems, leading to a valuable set of Radio Frequency Machine Learning (RFML) tools that can support both tactical-level RF situational awareness (SA) and strategic signals intelligence.
Many such RF applications are defined and leveraged in operational strategies highlighted in the Joint Chiefs' Electromagnetic Spectrum Operations (JEMSO) strategy, with RFML techniques offering user-selectable accelerations of real-time battlefield SA.\cite{jesmo}
The context for our research, and the coinciding goals of this paper, is to develop a foundational bases for quantifying and qualifying training data in supervised RFML Deep Learning (DL) systems to help designers balance cost, performance, and data types when constructing new RF situational awareness tools.

RFML has its roots in both commercial and military driven applications.\cite{5g}
One such product in the military domain that leverages the RFML foundation discussed here and extended from our research is SignalEye, which enables the war fighter on the tactical edge to have enhanced situational awareness of the battlefield by analyzing the RF spectrum automatically and detecting, isolating, and classifying signals.\cite{signaleye}
``Detection'' is the process of finding signals, which can be a few kilohertz wide within the gigahertz of spectrum, or have a duration on the scale of microseconds.
Such RFML-based detection algorithms can be tuned to specific signals of interest (SOI), or generalized to broader categories of {\it spectral triage} that focus on identifying spectral anomalies for further inspection.
While the process of ``classification'' can be perceived at different levels of fidelity, the classification space can be simplified in terms of Friend, Foe, and Other, enabling better awareness of direction and distance of emitted signals with enough relevant observations of the spectrum.
One specific way to increase the understanding for Friend/Foe determination is through the application of RF Fingerprinting, or more specifically Specific Emitter Identification (SEI), which is capable of identifying unique transmission devices in the case of adversarial spoofing of signals.\cite{wong-sei1,wong-sei2,sankhe2019oracle}
Therefore understanding how to create, train, and predict performance for systems that can best utilize these emerging RFML techniques is of significant interest.

For a given problem within the RFML space, once a reliable training routine and a network of sufficient size have been identified, how well a trained network is able to solve the problem often comes down to the quantity and quality of the data available \cite{Goodfellow-et-al-2016}.
Effectively, there are three sources of data that can be used to train networks within the RFML space: simulated/synthetic \cite{Nandi_1997,Kim_2003,Fehske_2005,Mody_2007,Ge_2008,Bixio_2009,adv_rfml:Clancy2009a,Ramon_2009,Popoola_2011,Kang_2011,Pu_2011b,He_2011,Abedlreheem_2012,Li_2012,Thilina_2013,Popoola_2013,Kim_2013,Tsakmalis_2014,Chen_2015,Mendis_2016,Oshea_2016,Oshea_2016c,oshea2016datagen,Oshea_2016d,west_2017,west-amc2,Karra_2017,oshea-detect2,Peng_2017,Reddy_2017,Nawaz_2017,Nawaz_2017b,Ambaw_2017,Ali_2017,Hong_2017,Yelalwar_2018,hauser-amc,Hiremath_2018,Tang_2018,adv_rfml:Davaslioglu2018a,Kulin_2018,Wu18,Li_2018,Tandiya_2018,Subekti_2018,Jayaweera_2018,oshea-2018-data,Zhang_2018,Sang_2018,Vanhoy_2018,Shapero_2018,Yashashwi_2019,Peng_2019,Wang_2019,Liu_2019,Zheng_2019,Wang_ARL_2019,Clark19, wong-sei2, merchant_2019b, Moore_2020}, captured/collected \cite{Mody_2007,Ge_2008,Cai_2010,Pu_2011b,Popoola_2013,ts_2014,Kumar_2016,Reddy_2017,Schmidt_2017,Vyas_2017,Bitar_2017,oshea-detect1,oshea-detect2,Kulin_2018,Fernandes_2018,Testi_2018,Yi_2018,merchant_2018,wong-sei1,Mohammed_2018,oshea-2018-data, Elbakly18,Mendis_2019,Hiremath_2019,sankhe2019oracle,AlHajri2019,Chawathe19, merchant_2019a}, and augmented \cite{Mody_2007,Schmidt_2017,adv_rfml:Davaslioglu2018a,merchant_2018,Kulin_2018,Wang_ARL_2019}, which is a combination of the first two using domain knowledge (focus of this work), or using Generative Adversarial Networks (GAN) as performed in Davaslioglu et al.\cite{adv_rfml:Davaslioglu2018a}
Due to the nature of the RFML data space, simulated data is inexpensive thanks to open source tool-kits like GNU Radio, where observations can be generated uniquely in parallel, with the only bottleneck being the available compute resources.\cite{oshea2016datagen}
Comparatively, performing an Over-the-Air (OTA) collection costs many orders of magnitude greater in terms of time and money due to procurement and configuration of the hardware transceivers and having to generate data in real time rather than in parallel as is done in simulation, yet all the work done in order to simulate the data is still needed when not directly examining Commercial Off-The-Shelf (COTS) equipment.
That cost only increases once collection is moved onsite to environments of interest for acquisition of the highest quality data, because robust mobile systems need to be assembled in order to perform collects while recording and storing the vast quantities of In-phase and Quadrature (IQ) data needed for training.
This is a familiar problem in other domains such as image processing, where labeled training datasets are augmented to expand the size of the datasets and improve neural network generalization performance \cite{Simard, NIPS20124824, NIPS19961250}, such that data augmentation becomes a viable alternative and builds off a comparably smaller collected dataset.
In Wang et al., using synthetic permutations of an Additive White Gaussian Noise (AWGN) class and combining that with other classes was enough to increase the performance of their network in the Army Rapid Capabilities Office's (RCO) Blind Signal Classification Competition with an augmentation factor of seven; i.e., adding seven augmentations per observation to the original dataset.\cite{Wang_ARL_2019}

In order to understand the impact augmentation brings to RFML, an application space is needed without loss of generalization.
Due to the widely studied signal classification problem of Automatic Modulation Classification (AMC) within RFML, the AMC problem space makes a good way to test the promise of augmentation in RFML data without having to perform a full exploratory study determining the network and training routines needed in order to perform well.
Work performed by Sankhe et al. showed that in the RFML category of RF Fingerprinting that when the channel, or propagation path, changes and the new channel is not in the training data, the performance of the RFML system rapidly degrades.\cite{sankhe2019oracle}
From that it can be concluded that the quality of a dataset will be higher if the propagation path of the intended environment for field usage is within the dataset.
In simpler terms, the training data is known to be better when collected from the same physical location and conditions as what the deployed system will encounter.
Additionally, not only does the environment matter, but the role of the algorithms in detecting and isolating a signal also play an important role since these imperfections in the algorithms have an impact on a network's performance in AMC when not considered.\cite{hauser-amc}
One final factor known for AMC is that for diverse waveform spaces, a significant amount, i.e., $>1M$ observations, of data is required.\cite{oshea-2018-data}
In order to better gauge augmentation, focus will be on augmenting the detection imperfection space, Frequency Offset (FO) and Sample Rate Mismatch (SRM), along with varying the effective Signal-to-Noise Ratio (SNR), as these are computationally cheap augmentations rather than trying to quantify the imperfections observed in the propagation path.
Here, propagation path refers to any effects that deviate the signal from ideal digital representation, which include everything from the transmitter's Digital-to-Analog Converter (DAC) up to the receiver's Analog-to-Digital Converter (ADC).
The detection imperfections are taken as a post processing effect imposed by the detection algorithm after the receiver's ADC.
While the work in O'Shea et al.\cite{oshea-2018-data} does perform AMC with OTA data collection, it does so in a relatively benign environment, therefore the work is closer to what Sankhe et al.\cite{sankhe2019oracle} called a {\it static} channel, rather than a more realistic {\it dynamic} channel to which significantly more preprocessing was used to overcome.

This work seeks to investigate several questions open in the field.
First, with no first hand knowledge of the degradation of the signals to be seen, how well does a synthetically trained, validated, and tested network actually perform under real-world conditions in the field?
The major investigation here is the contrast between a synthetic dataset where, through simulation, distortion is applied to the signals and compared to the field collection of data where all of the distortion is taken from the signals' propagation through the environment.

Second, what value does augmentation bring in contrast to just performing an extended capture?
This question addresses the initial data collection concerns when starting a new problem or a repeat application within a new environment, which is because deep learning typically has a nonlinear relationship between performance and the number of observations in a dataset.
For narrowband signals, getting another order of magnitude of examples may change the length of time running a collection from days into months of field time.
In the absence of enough data, augmentation is relied upon to provide a greater observational data space, and by comparison, is relatively cheap in terms of time and effort to that of a traditional collection campaign; however, the value of one augmented observation contrasted with one collected observation in terms of performance is not well understood.
Further, many military spectrum access applications require modeling of channel effects that cannot be practically tested live (e.g., electronic warfare or atmospheric scintillation).

\begin{figure*}[ht]
    \includegraphics[width=\textwidth]{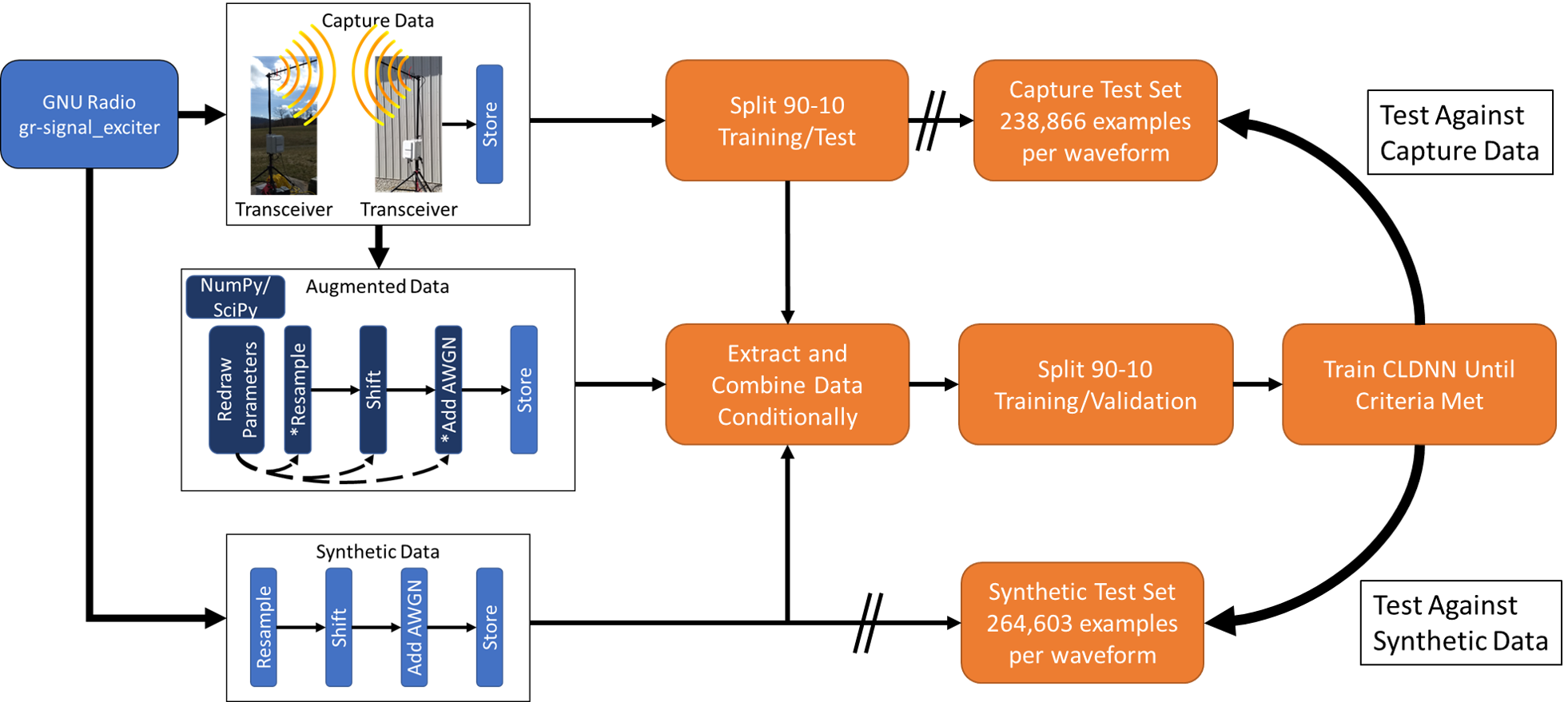}
    \caption{Experiment Configuration. At the start of each experiment, data is extracted from either a synthetic database, a capture database, and/or conditionally from an augmented database based on the extraction from the capture database. All data is then combined and split into 90\%/10\% training/validation datasets used to train a CLDNN network until an exit condition is met. The `//' indicate a separate file has been created to better isolate the testing data from the training data.}
    \label{fig:sysdiagram}
\end{figure*}

The third question investigated here is whether understanding the distributions of degradation sources impacts the ability of a network to achieve peak performance.
That is, can an approximation made by an expert over a narrow range be sufficient to allow the network to generalize over the full degradation space, or will taking measurements of the degradation space and utilizing those measured values prove to be more beneficial to the network's performance in that measured range?
There are two cases to examine under this question: synthetic generation and augmentation.
For the synthetic portion of this inquiry, the focus is on whether drawing parameters from the assumed degradation region, or drawing parameters from estimations extracted from the observation space allow for any change in performance of the trained network when tested against collected data.
With the augmentation portion, the question is whether it matters how the degradation space is resampled in the augmented observations; for example, can augmentation be performed using an assumed parameter degradation region like with the synthetic example, or should the resampling come from the collected observation space instead.

The work in this paper reinforces the results of Hauser et al., which show the effects of the detection and isolation algorithms should not be ignored for practical real-world applications, and when used intelligently with augmentation can improve performance.\cite{hauser-amc}
However, as shown in the work of Sankhe et al., the propagation path is a greater barrier to high fidelity synthetic data than the detection and isolation imperfections alone.\cite{sankhe2019oracle}
The quantification of the training data required to achieve a desired level of performance in a DL RFML application will support predictive measures of data needs for other applications.
These results match an intuitive expectation that marginal performance will asymptotically approach a fundamental maximum based on the quality of the training data, regardless of the quantity of data available.

The rest of this paper is outlined as follows.
In Section \ref{sect:experiment}, a description of the experiment performed in this work is outlined and covers the model architecture, training routine, database used, and the AMC problem spaces being trained.
In total, three AMC problem spaces are considered for a contrast of application difficulty, each with a different number of waveforms present in the space, consisting of 3, 5, and 10 waveforms.
In Section \ref{sect:results}, the primary results from the three waveform spaces are presented and discussed in terms of the three questions outlined above, focusing on understanding how the quantity and quality of a dataset impact AMC performance.
Section \ref{sect:future} poses open questions that this work raises for future efforts and broadens consideration to other RFML applications.
And finally, in Section \ref{sect:conclusion}, parting thoughts and conclusion are provided.

\begin{figure*}[ht]
    \includegraphics[width=\textwidth,trim=45 40 0 5,clip]{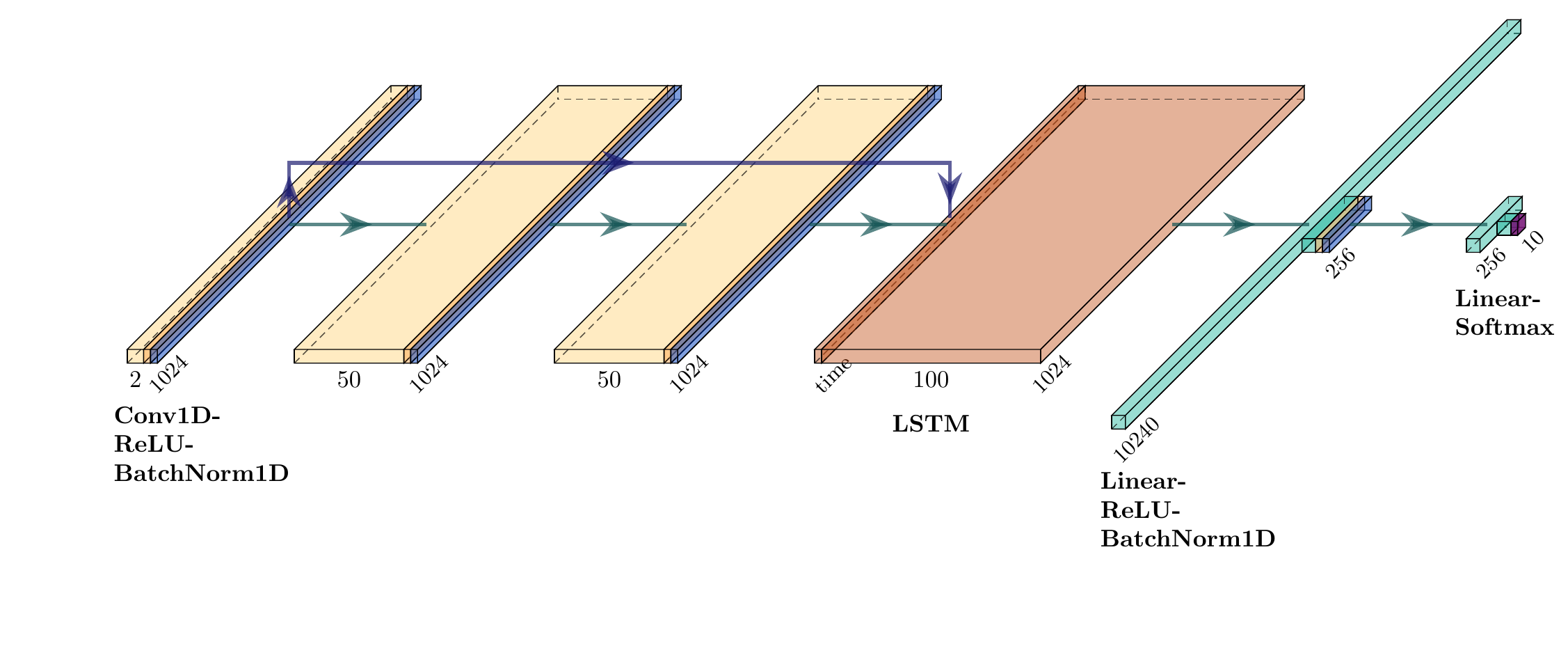}
    \caption{CLDNN Architecture for the $\Phi_{10}$ (Discussed in Section \ref{sect:waveformspace}) waveform dataset.}
    \label{fig:arch}
\end{figure*}

\section{Experiment Setup}\label{sect:experiment}

As the focus of this work is to examine the effects of quantity and quality of the available data to the RF problem space of AMC, the architecture for the model to be trained along with a DL training routine were identified and remain consistent for all aspects of the results shown within this work.
The experiment, shown in Figure \ref{fig:sysdiagram}, consists of training a Convolutional, Long-Short Term Memory (LSTM) Deep Neural Network (CLDNN) for a maximum of 50 epochs through all available training data after a 90\%/10\% split for training and validation datasets respectively.
For the case where there are 101 observations for a waveform, 91 observations per waveform are in the training dataset while 10 per waveform are in the validation dataset.
A second condition for stopping is allowed for in the form of early exiting when the validation loss does not decrease for 4 epochs, and is responsible for all results presented here as the maximum number of epochs present is 34.
A similar training routine and process were used in our prior work, with multiple such examples achieving operational deployment.\cite{hauser-amc,signaleye}
A discussion for the selection of the architecture, training routine, and datasets follow.

\subsection{Network Architecture}\label{sect:ntwkarch}
An extensive investigation of RFML architecture was under taken by West et al., where different networks from the literature were compared and contrasted.\cite{west-amc2}
The dominant network for performing AMC was found as the CLDNN, and a practical description for implementation was given by Flowers et al. with the addition of batch normalization throughout the network.\cite{rfml2019tutorial}
From these works, the network used in this experiment, as seen in Figure \ref{fig:arch}, is then a CLDNN, which accepts an input of two channels (In-phase and Quadrature floats) with 1024 samples passing through three 1D convolution layers with 50 output channels, using a 1x8 kernel size, whose input is padded with zeros such that the output sequence is of the same length as the input, followed by a Rectified-Linear Unit (ReLU) activation, and a 1D batch normalization layer.
The output of the first and third such layers are then concatenated along the channel dimension and passed through a single LSTM layer such that the channels are taken as the features while the time sequence is fed through the LSTM's memory structure.
For simplicity, the LSTM has a hidden size equal to the number of classes being used in the problem.
The output of the LSTM is then flattened and passed through a linear layer with 256 output nodes with a ReLU Activation and 1D Batch Normalization.
The final batch normalization layer is then connected to the final linear layer with the number of outputs equal to the number of classes with a Softmax Activation.
This network structure was chosen as the baseline for the experiment for two main reasons, the first being the high performance seen in West et al.\cite{west-amc2}, and the second being that the convergence time with the training routine given next was shown to be quick in terms of epochs by Flowers et al.\cite{rfml2019tutorial}

\subsection{Training Routine}\label{sect:trainingroutine}
The CLDNN network is trained consistently for all data quantities and qualities.
As discussed next, the dataset for each experiment is selected, and then split 90\%/10\% into training and validation datasets respectively.
These two datasets are then distinct for that experiment and model trained.
The data is then processed in batches consisting of 1500 total observations.
Using the default Adam optimizer \cite{adam_opt} within PyTorch\cite{pytorch} with Cross Entropy Loss, the network is allowed to continually train until 4 epochs have passed without any improvement in the validation loss metric or until a total of 50 epochs have been processed.
The experiment keeps the weights whose validation was the least across all trained epochs for the model because as the validation rises after such a minimum is reached, the training routine assumes the weights have begun overfitting.

While this training routine doesn't allow for significant deviation in the event a local minimum is found, it is chosen for the ideal properties of having a quick consistent goal and a limited processing window within which to achieve said goal.
In order to best understand the relationship between performance and quantity of available data, hundreds to thousands of networks need to be trained in order to regress the relationship while accounting for unknown outlier potentials; allowing a more lenient training routine would only extend the total time taken to train a single model.
The approach above was selected while taking into consideration the total processing time needed to perform the investigation.

\subsection{The Dataset}\label{sect:thedataset}
The database being used in this investigation was made from live collection of numerous waveforms at Virginia Tech's Kentland Farms over the course of 4 months in 2018 using two weather enclosed software-defined radios (shown in Figure \ref{fig:sysdiagram}).
The database consists of multiple waveforms of varying duration and quality.
Based on the metadata available, the data was filtered to only select observations whose SNR was estimated to be above -10 dB.
Additionally, due to the nature of hardware collections and an active environment, observations that were found to have irregular sample values were filtered out from the dataset.
The data was then segmented such that observations of 1024 samples could be extracted in a continuous fashion with no two observations being contiguous outside of 1024 samples.
In other terms, given one observation starting at sample 0, the next observation could not start until sample 2048 assuming there are at least 3072 samples available in that record.
The final filter placed on the data left all waveforms evenly balanced in terms of available observation counts, resulting in 2,388,667 total observations for each waveform class considered.
Of the data that remained, the detector imperfections that were estimated showed that the SNR was between -10 and 80 dB, with the majority of the data being below 20dB.
The estimated FO were found to be bounded by $\pm$20\% of the receiver's sampling rate though heavily concentrated between $\pm$5\%.
The SRM found signals in the range of 2-32 times that of the Nyquist rate, though roughly twice as likely to be between 2-8 as between 8-32.
In order to better make use of the estimated distributions of the captured data, a joint Kernel Density Estimate (KDE) was performed per modulation on the imperfections of SNR, FO, and SRM to be used while augmenting and synthetically generating datasets.
Table \ref{tab:datasets} provides an overview for the datasets used in this work.

\begin{table}
\small\sf\centering
\caption{Description of datasets used within this work.}\label{tab:datasets}
\begin{tabular}{c c l}
\toprule
\multicolumn{3}{c}{{\bf Training}}\\
\midrule
Symbol & Source & Description\\
\midrule
\multirow{2}{*}{$\Omega_{C}$} & \multirow{2}{*}{Capture} & Consists of only capture\\
& & examples\\
\midrule
\multirow{4}{*}{$\Omega_{SS}$} & \multirow{4}{*}{Synthetic} & Consists of simulated\\
& & examples using\\
& & an assumed synthetic\\
& & distribution\\
\midrule
\multirow{5}{*}{$\Omega_{AS}$} &  & Consists of capture\\
& Capture & examples and\\
& + Synth & augmentations\\
& Augmentation & using the synthetic\\
& & distributions\\
\midrule
\multirow{3}{*}{$\Omega_{SK}$} & \multirow{3}{*}{\begin{tabular}{@{}c@{}}Synthetic \\ using KDE\end{tabular}} & Consists of simulated\\
& & examples using the KDE\\
& & of the capture dataset\\
\midrule
\multirow{5}{*}{$\Omega_{AK}$} & & Consists of capture\\
& Capture & examples and\\
& + KDE & augmentations\\
& Augmentation & using the KDE\\
& & of the capture dataset\\
\midrule
\multicolumn{3}{c}{{\bf Testing}}\\
\midrule
\multirow{2}{*}{$\Omega_{TC}$} & \multirow{2}{*}{Capture} & Consists of only capture\\
& & examples\\
\midrule
\multirow{4}{*}{$\Omega_{TS}$} & \multirow{4}{*}{Synthetic} & Consists of simulated\\
& & examples using\\
& & an assumed synthetic\\
& &distribution\\
\bottomrule
\end{tabular}
\end{table}

\subsubsection{The Capture Dataset}\label{sect:thecapturedataset}
All the observations in the dataset described above are then split into a general training set ($\Omega_{C}$) and a test set ($\Omega_{TC}$) against which all results will be compared.
The training/testing split is 90\%/10\% consisting of 2,149,801 and 238,866 observations per class respectively.
As part of the investigation, the number of observations drawn from the training set varies across iterations, but every trained model is tested against the full $\Omega_{TC}$.

\subsubsection{The Augmented Dataset}\label{sect:theaugmenteddataset}
Additionally, there are other datasets that can be used while training consisting of captured data.
The first is an augmented dataset that is linked to the capture training set and, for every observation, 10 augmentations are made.
The data drawn from the augmented dataset is conditionally linked in such a way that only augmentations of data selected from the capture training dataset will be available to be drawn, and then when the augmentation factor is less than 10, which exact augmentation is drawn is left to random uniform sampling.
There are two distinct augmented datasets to understand what effect, if any, the distribution of parameters has on the performance during training.

The first augmented dataset takes on a range of parameters given as an expected performance range of the capture data prior to performing any capture.
The signals are augmented in such a way that the SNR is uniformly drawn from the range of 0-20dB, the FO is taken uniformly in the range of $\pm$10\% of the sampling rate, and the SRM is taken to be uniform in the range of 2-8 times that of the Nyquist rate for the captured signals.
Given the available metadata and the observations of 1024 samples, any time a random value is drawn that cannot be achieved, for example a signal with an estimated SNR value of 5dB attempting to augment the signal to an SNR of 10dB, the augmentation is nulled and whatever the current estimate is holds.
Likewise, SRM augmentation that requires decimation reducing the number of samples below the desired 1024 observation length is nulled.
This dataset is the $\Omega_{AS}$ dataset, and the parameter space is drawn from three independent distributions using NumPy.\cite{numpy}

The second augmentation dataset makes use of a Gaussian joint kernel density estimate, using SciPy, of the available data per captured waveform and uses a random draw from that joint estimate to perform the augmentation.\cite{scipy}
This dataset is the $\Omega_{AK}$ dataset.
In this way, the investigation can contrast the value of data analysis on the captured data with regard to augmentation or whether a general blind practical range will suffice.

\subsubsection{The Synthetic Dataset}\label{sect:thesyntheticdataset}
The final two datasets consist of simulated synthetic data for the waveforms under test.
Using the same distribution assumptions for the SNR, FO, and SRM as the $\Omega_{AS}$ dataset, the synthetic dataset randomly generates observations for each waveform to be used during training, $\Omega_{SS}$.
A second synthetic training set is used under the assumption that better metrics are known for SNR, FO, and SRM based on the targeted detection routine in place on the observer device and draws the parameters from the KDE discussed with the $\Omega_{AK}$ dataset; the $\Omega_{SK}$ dataset.
This approach can help quantify the value of real world data that undergoes true transceiver and channel degradation that is not as easily replicated through simulation.
Additionally, a single testing dataset, $\Omega_{TS}$, is created using the blind distributions as a means for comparing what a purely synthetic test set would say about the performance of a trained model to that of captured data from the field.

\subsection{Waveform Space}\label{sect:waveformspace}
\begin{table}
\small\sf\centering
\caption{Waveform Spaces. Different waveforms used to examine the problem space as the complexity changes.}\label{tab:waveform_space}
\begin{tabular}{c c}
\toprule
Group & Waveforms\\
\midrule
$\Phi_{3}$ & BPSK, QPSK, Noise\\
\midrule
$\Phi_{5}$ & BPSK, QPSK, QAM16, QAM64, Noise\\
\midrule
\multirow{2}{*}{$\Phi_{10}$} & BPSK, QPSK, QAM16, QAM64, BFSK,\\
& GMSK, AM-DSB, FM-NB, GBFSK, Noise\\
\bottomrule
\end{tabular}
\end{table}

The waveforms selected from those available in the data consist of 3, 5, and 10 classes denoted by $\Phi_{3}, \Phi_{5},$ and $\Phi_{10}$, respectively.
The waveforms associated with each space are provided in Table \ref{tab:waveform_space}.
By having three different dimensions for the class size, the work is able to examine any differences that data quantity might have with regard to the difficulty of the problem.
The two smaller subsets are chosen due to the frequent usage in traditional feature based AMC approaches.\cite{swami_2000}

\section{Analysis}\label{sect:results}
The results presented in this section are aimed at investigating each of the three research questions posed previously.
In doing so, the overall analyses employ Monte Carlo ensembles that consider relative performance against synthetic ($\Omega_{TS}$) and captured ($\Omega_{TC}$) test datasets, and ultimately analysis of AMC application accuracy as a function of the quality and quantity of data provided during training.

\subsection{Synthetic Performance in the Field}\label{sect:synthperf}
\begin{figure*}[t]
    \includegraphics[width=\textwidth, trim=145 10 160 56, clip]{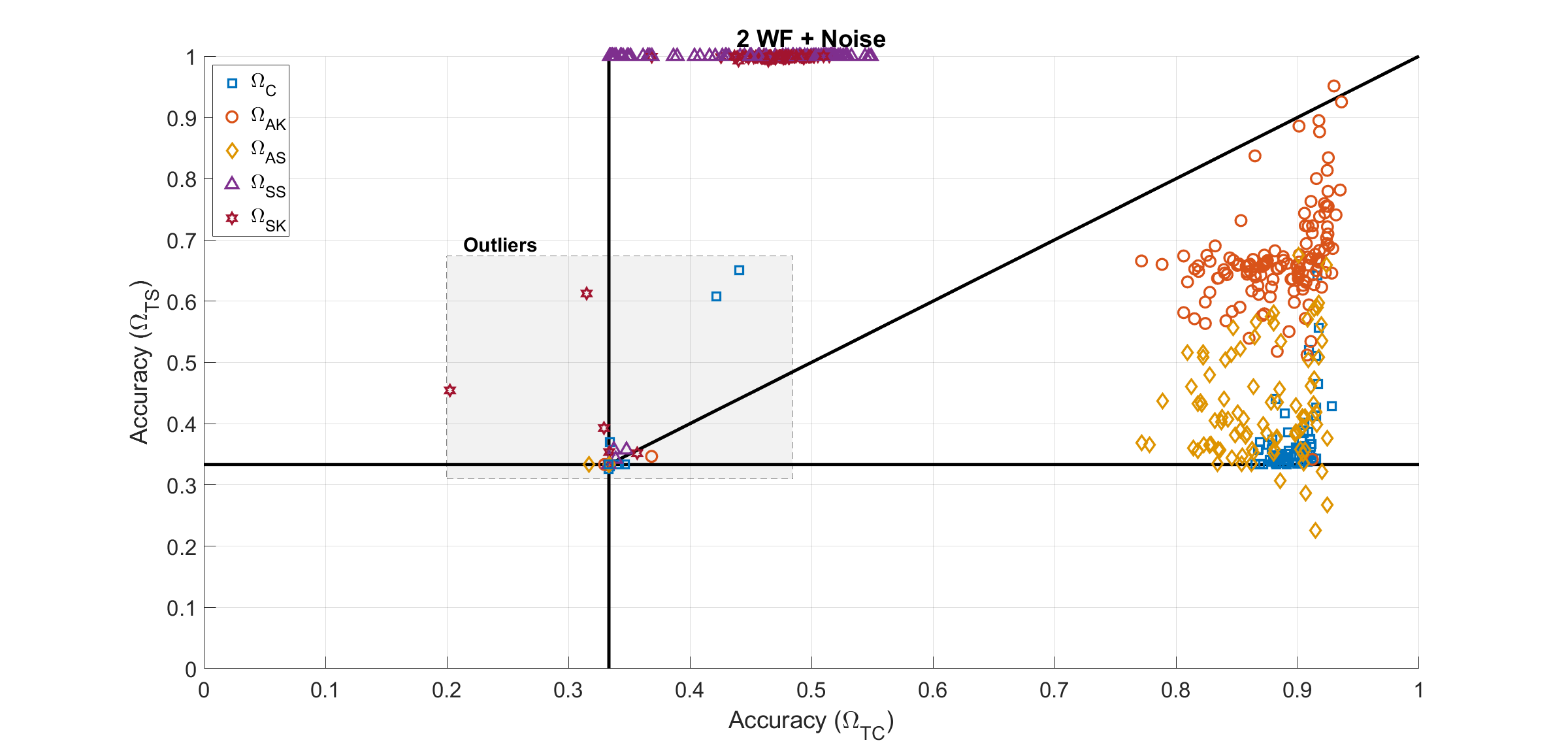}
    \caption{Performance of individually trained networks using the five datasets on the $\Phi_{3}$ waveform space contrasting the performance observed on the synthetic ($\Omega_{TS}$) and capture ($\Omega_{TC}$) test sets. While ideal performance is in the top right corner, an acceptable performance is to the right in general for real-world performance. While high performance is found on data similar to their respective training datasets, neither synthetic nor captured data performs well on the other. Through augmentation of captured, a greater performance is observed on the synthetic data as a result indicating better generalization over the full parameter space.}
    \label{fig:perf_compare_03}
\end{figure*}

\begin{figure*}[t]
    \includegraphics[width=\textwidth,trim=145 10 160 56, clip]{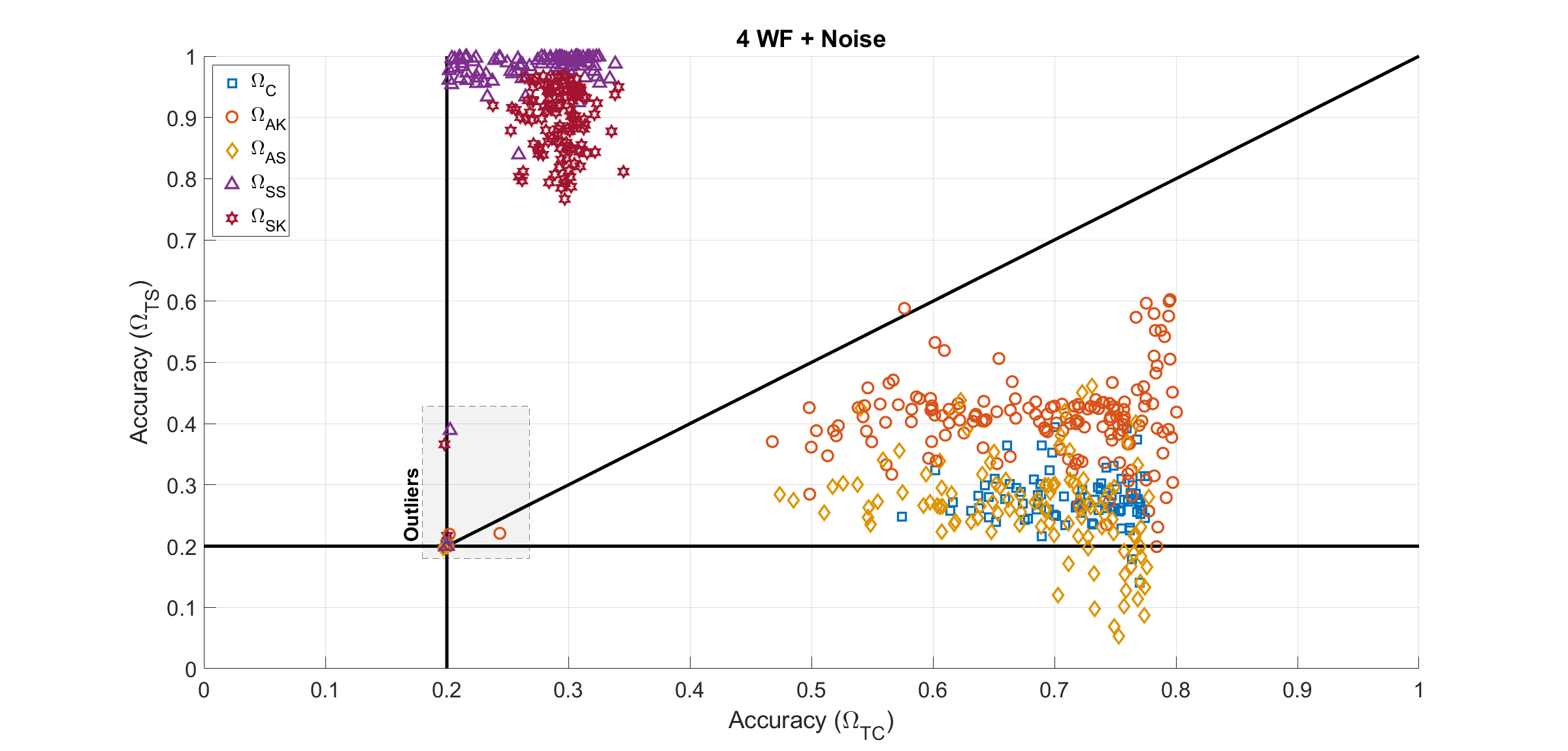}
    \caption{Performance of individually trained networks using the five datasets on the $\Phi_{5}$ waveform space contrasting the performance observed on the synthetic ($\Omega_{TS}$) and capture ($\Omega_{TC}$) test sets. While ideal performance is in the top right corner, an acceptable performance is to the right in general for real-world performance. In contrast to Figure \ref{fig:perf_compare_03} the generalization over the parameter space is less pronounced when using augmented data in the same quantity range per waveform; however, the augmented dataset with knowledge of the parameter space ($\Omega_{AK}$) still provides improved generalization, while augmentation without such knowledge ends up reducing generalization instead.}
    \label{fig:perf_compare_05}
    
    \cleardoublepage
\end{figure*}

\begin{figure*}[t]
    \includegraphics[width=\textwidth,trim=145 10 160 56, clip]{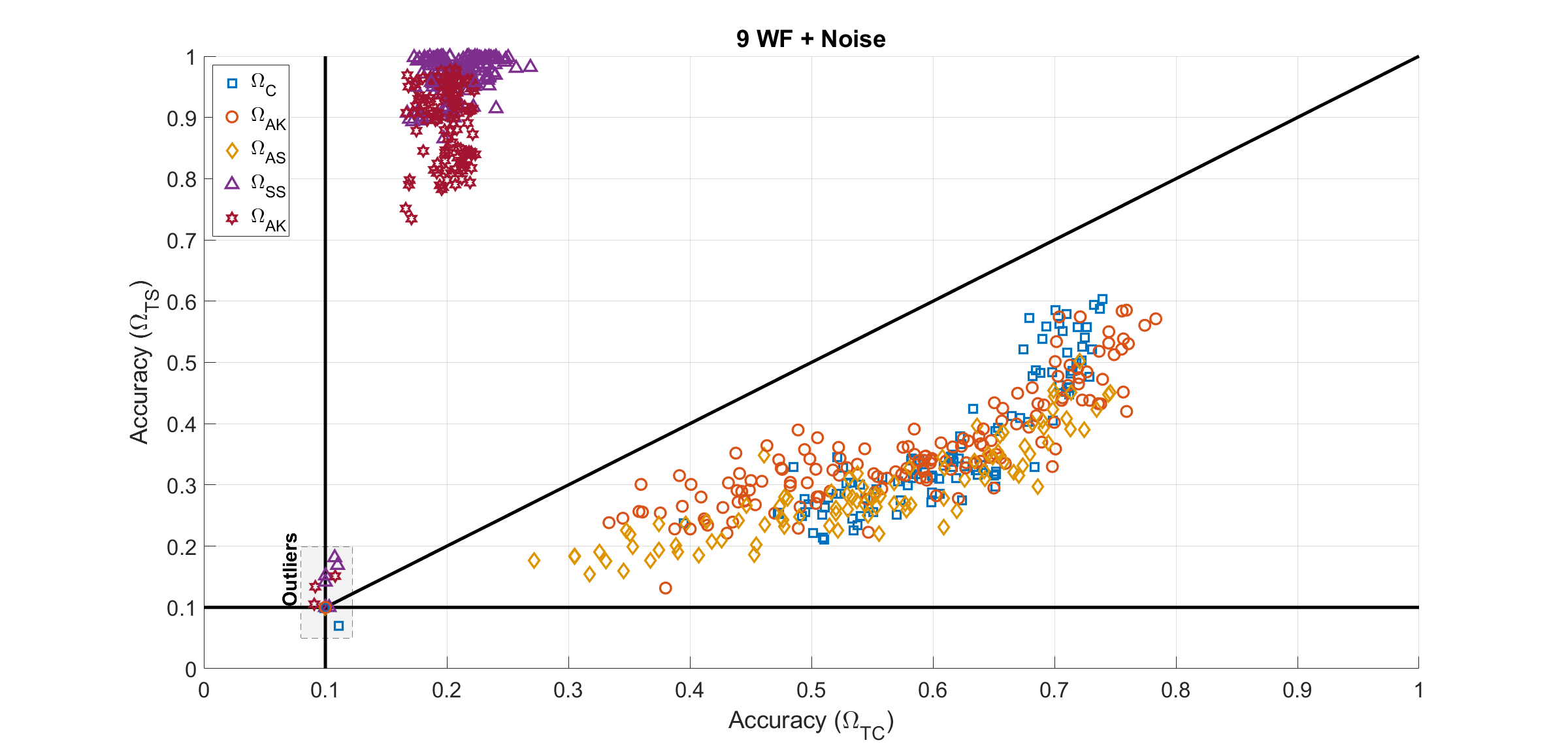}
    \caption{Performance of individually trained networks using the five datasets on the $\Phi_{10}$ waveform space contrasting the performance observed on the synthetic ($\Omega_{TS}$) and capture ($\Omega_{TC}$) test sets. While ideal performance is in the top right corner, an acceptable performance is to the right in general for real-world performance. In contrast to Figures \ref{fig:perf_compare_03} and \ref{fig:perf_compare_05} the generalization observed through augmentation is only observable at low ac curacies on $\Omega_{TC}$ and becomes negligible as accuracy increases. As the difficulty increases, the improved generalization that can be observed with augmented data within lower difficulty problems disappears in higher difficulty problems when constrained to the same quantity/quality availability during training. It is unclear whether quantity or quality directly play a role of significance in this observation.}
    \label{fig:perf_compare_10}
\end{figure*}

The analysis starts with the first major question in this work; for a given synthetically trained network, how well does the network perform when applied to real-world data from the field?
To answer this question, three different waveform spaces, $\{\Phi_{3},\Phi_{5},\Phi_{10}\}$, are examined with regard to the dataset source used while undergoing training.
Each trained model is then evaluated on both the synthetic and the capture test sets, $\{\Omega_{TS}, \Omega_{TC}\}$.
Plotting the accuracy of the $\Omega_{TC}$ against the accuracy of the $\Omega_{TS}$ in Figures \ref{fig:perf_compare_03}-\ref{fig:perf_compare_10}, the overall performance can be seen for the waveform spaces $\{\Phi_{3},\Phi_{5},\Phi_{10}\}$ respectively.
For each plot, a vertical and horizontal black solid line, without markers, are used to indicate random guessing along each axis, with a third diagonal line indicating where performance would be equivalent across the two test sets.
This diagonal line is clipped where it meets the first two lines because falling closer toward zero with any significance would require intentional or adversarial manipulation of the training routine or data, which is not considered in this work.\cite{flowers_adv}
The trained networks are then represented with different markers representing different dataset sources used during training.


In terms of an ideal performance, non-filled markers should be concentrated in the top right corner of each plot, indicating high accuracy on both the $\Omega_{TC}$ and $\Omega_{TS}$ datasets, which can be used as an indication that the nuisance parameter space in the capture data that is not being modeled in the synthetic data has been well generalized over.
Instead of this ideal performance, two distinct cases are seen, higher performance on $\Omega_{TS}$, or higher performance on $\Omega_{TC}$.
The data types that performed best on the synthetic test set were the models trained from the synthetic training datasets $\Omega_{SS}$ and $\Omega_{SK}$, using marker shapes triangle and pentagram respectively.
Whereas the data types that performed best on the capture test set were the models trained from the capture and augmented datasets $\Omega_{C}$, $\Omega_{AK}$, and $\Omega_{AS}$, using marker shapes square, circle and diamond respectively.

While this bias for alike datasets intuitively makes sense, there are some unique outcomes that are not obvious.
There are grayed out areas marked as outliers, which are results that failed to converge to a performance using the constraint
\begin{equation}
    \text{Lower Bound} = \max\left(\alpha_{\Omega_{C||S}}\right) \mid \alpha_{\Omega_{C||S}} < \frac{2}{|\Phi_{X}|}.
\end{equation}
Here any accuracy, $\alpha$, that is less than or equal to the lower bound is treated as an outlier and excluded from the analyses within this paper to allow for focus to be on the convergent networks, while $X$ is used as the waveform space indicator.
This filter corresponds to models that did not perform better than $[44.05\%, 24.36\%, 11.13\%]$ from datasets \{$\Omega_{C},\Omega_{AK},\Omega_{AS}$\} and $[61.25\%,38.91\%,18.14\%]$ from datasets \{$\Omega_{SS},\Omega_{SK}$\} set in the waveform spaces $\{\Phi_{3},\Phi_{5},\Phi_{10}\}$ respectively.

\begin{table}
\small\sf\centering
\caption{Test Accuracy's Observed Response to Synthetic Datasets $\Omega_{SS}$ and $\Omega_{SK}$. Examines the significance of where the detection imperfections are drawn from compared to what is tested against. Smaller values ($<0.05$) indicate larger significance of where the simulation degradation is drawn from. p-values found using Welch's two sample t-test.}\label{tab:synth_p_values}
\begin{tabular}{c c c c c}
\toprule
\multirow{4}{*}{\begin{tabular}{@{}c@{}}Waveform \\ Space\end{tabular}} & \multicolumn{2}{c}{Average} & \multicolumn{2}{c}{\multirow{3}{*}{p-values}}\\
 & \multicolumn{2}{c}{Accuracy Ratio} & & \\
 & \multicolumn{2}{c}{$\overline{\Omega_{SK}}/\overline{\Omega_{SS}}$} & & \\
 & $\Omega_{TS}$ & $\Omega_{TC}$ & $\Omega_{TS}$ & $\Omega_{TC}$\\
\midrule
$\Phi_{3}$ & $0.998$ & $1.05$ & $1.2196\mathrm{e}{-10}$ & $3.9351\mathrm{e}{-3}$ \\
\midrule
$\Phi_{5}$ & $0.910$ & $1.08$ & $1.3715\mathrm{e}{-35}$ & $5.6249\mathrm{e}{-6}$ \\
\midrule
$\Phi_{10}$ & $0.921$ & $0.954$ & $1.9963\mathrm{e}{-23}$ & $2.6340\mathrm{e}{-5}$\\
\bottomrule
\end{tabular}
\end{table}

Observing the figures for the synthetic trained datasets ($\Omega_{SS}$: triangle and $\Omega_{SK}$: pentagram markers) on the performance comparison show that the synthetic data is very easily learned when tested on $\Omega_{TS}$, but generally fails to do better than twice that of random guessing on $\Omega_{TC}$.
Table \ref{tab:synth_p_values} shows the relational change in performance on both $\Omega_{TS}$ and $\Omega_{TC}$ when contrasting the two synthetic datasets.
Contrasting the average accuracy as a ratio, $\overline{\alpha_{\Omega_{SK}}}/\overline{\alpha_{\Omega_{SS}}}$, there is a marginal loss in performance on the smallest waveform space, $\Phi_{3}$, when using the synthetic data created with the KDE, while the KDE drawn synthetic data performs a little better on average with $\Omega_{TC}$ than the non-KDE drawn synthetic data; using Welch's two-sample t-test, the significance of such improvement is minimal.\cite{welch1947generalization}
By contrast, the decrease in average accuracy on $\Omega_{TS}$ and the increase in average accuracy on $\Omega_{TC}$ for $\Phi_{5}$ show a much greater significance indicating that the use of the imperfections from the intended environment can improve the fidelity of synthetic datasets.
However, the improvement on the capture test set is lost as the waveform space continues to grow with $\Phi_{10}$.
In general, there is a slight possibility that creating synthetic data that only considers the detector imperfections can be of high enough fidelity to train as the number of synthetic examples increases by many order of magnitudes, but overall these results show that modeling only the detector imperfections while ignoring the propagation path is not significant enough to properly train a system heading to the field.
This result answers the first question of when given a network trained and tested in the synthetic space, that network will not perform well in a real system without a much higher fidelity simulated dataset.
Additional work is still needed to determine at what threshold simulated data can be considered high enough fidelity when designing and developing a deployable system.
In all likelihood, finding that threshold is going to be very dependent on the target operating environment.
This includes how much is known about the transceiver-to-transceiver propagation path, which includes everything from the transmitter's DAC through the receiver's ADC, and any effects of the detection and isolation stages inherent to that receiver.

\begin{table}
\small\sf\centering
\caption{Log-Linear fits, $\text{qty}=10^{\left(\frac{\alpha-p_2}{p_1}\right)}$, for data presented in Figures \ref{fig:perf_capt_qty_03}-\ref{fig:perf_capt_qty_10}.}\label{tab:fits}
\begin{tabular}{c c c}
\toprule
\multirow{2}{*}{Dataset} & \multicolumn{2}{c}{Waveform Space $(p_1,p_2)$}\\
 & $\Phi_{3}$ & $\Phi_{5}$ \\
\midrule
$\Omega_{C}$ &  $0.03237,0.7485$ & $0.09351,0.2995$\\
\midrule
$\Omega_{AK}$ & $0.05091,0.6317$ & $0.1138,0.1402$\\
\midrule
$\Omega_{AS}$ & $0.05476,0.5955$ & $0.1022,0.1686$\\
\midrule
$\Omega_{SS}$ & $0.04183,0.2656$ & $0.01537,0.2030$\\
\midrule
$\Omega_{SK}$ & $0.002380,0.4650$ & $-0.002029,0.3019$\\
\midrule
& \multicolumn{2}{c}{$\Phi_{10}$}\\
\midrule
$\Omega_{C}$ & \multicolumn{2}{c}{$0.1459,-0.01837$}\\
\midrule
$\Omega_{AK}$ & \multicolumn{2}{c}{$0.1540,-0.1294$}\\
\midrule
$\Omega_{AS}$ & \multicolumn{2}{c}{$0.1598,-0.2043$}\\
\midrule
$\Omega_{SS}$ & \multicolumn{2}{c}{$0.008621,0.1721$}\\
\midrule
$\Omega_{SK}$ & \multicolumn{2}{c}{$-0.001438,0.2050$}\\
\bottomrule
\end{tabular}
\end{table}

\subsection{Value of Augmentation}\label{sect:valueaugment}
To start answering the second question of what value does augmentation bring to the problem, the attention shifts focus to the proximity of the models trained with $\Omega_{C}$, $\Omega_{AK}$, and $\Omega_{AS}$ datasets (square, circle, and diamond markers, respectively) to that of the diagonal line where there are performance generalizations that become less pronounced as the waveform space grows.
For the capture dataset models, the clusters show that $\Omega_{AK}$ typically achieves better performance on $\Omega_{TS}$ than both $\Omega_{C}$ and $\Omega_{AS}$, indicating that the degradation encountered from imperfect detector estimation when accounted for in augmentation, does help the network better generalize over the nuisance parameters present in the capture data.
Conversely, and more surprisingly, augmenting the dataset with the assumed synthetic range actually made the performance on the $\Omega_{TS}$ worse than without the augmentation.
One conclusion that can be drawn from this is that the degradation encountered between one transceiver's DAC to another transceiver's ADC has a greater effect on performance than the degradation caused by the detection algorithm's imperfections, assuming detection and isolation is achieved, and that simply redrawing the parameters occurred by one detection routine for another detection routine will not be sufficient without taking into account the propagation degradation on the path between the DAC and ADC in this new environment.
Such refinements will become more important as RFML systems begin to incorporate learned behaviors that have been trained with and transferred from another node.

\begin{figure*}[t]
    \includegraphics[width=\textwidth,trim=0 0 0 0, clip]{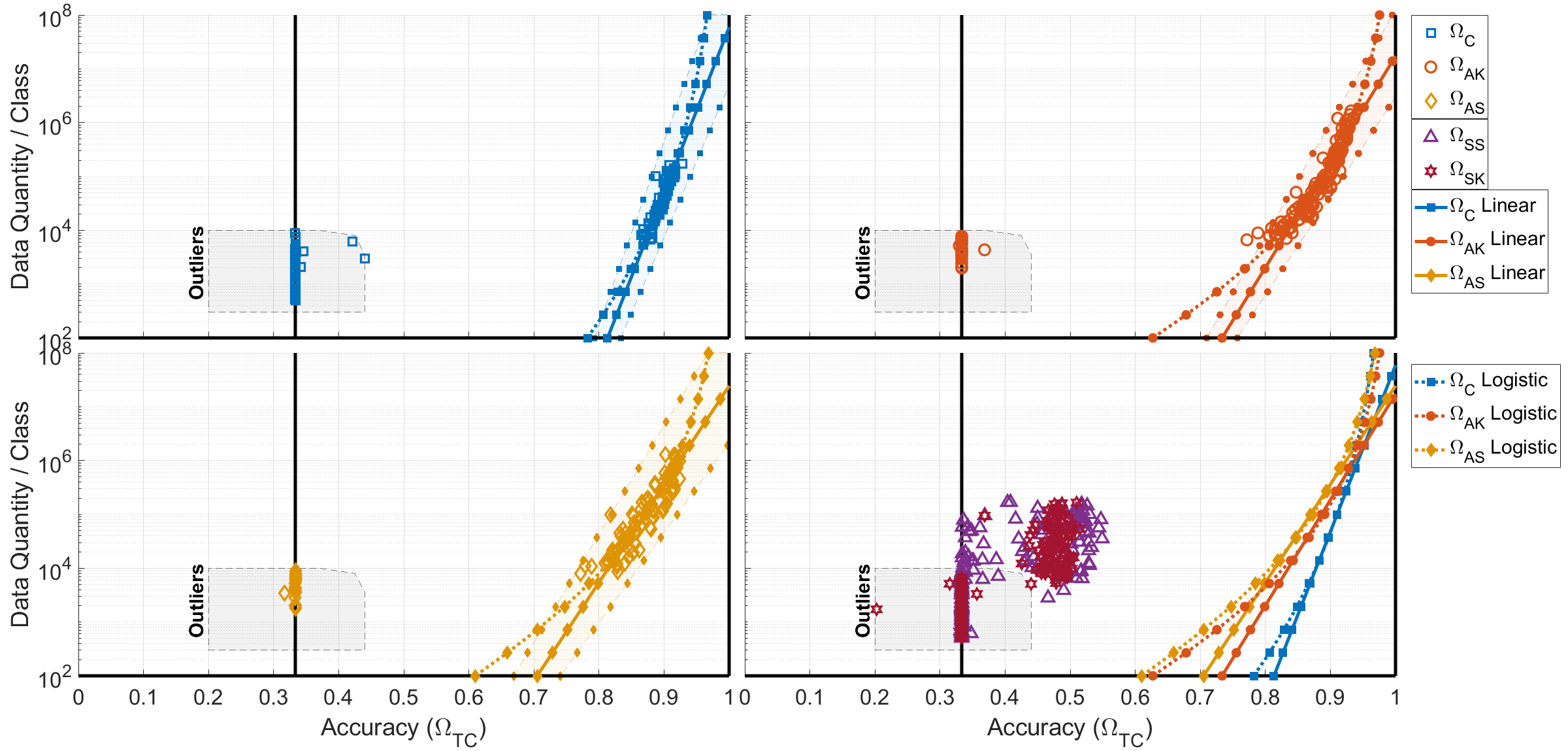}
    \caption{Performance of models trained using the five datasets on the $\Phi_{3}$ waveform space. The solid vertical line represents a network that is performing as well as a random guess. Solid lines with markers represent the trend in terms of examples per class that are needed to achieve a given accuracy on the capture test data, while the shaded regions between dashed lines with the matching markers indicate a 95\% confidence region for that trend. Synthetic datasets are omitted from the trend analysis as no significant trend was observed from these datasets. Trends are derived with outliers removed. The further right the trend line, the higher the quality of data. Additionally, a dotted line is fit to the data using a logistic regression with the assumption that 100\% accuracy is possible given the asymptotic curve perceived in the data and the impossibility of performance greater than 100\% accuracy.}
    \label{fig:perf_capt_qty_03}
\end{figure*}

\begin{figure*}[t]
    \includegraphics[width=\textwidth,trim=0 0 0 0, clip]{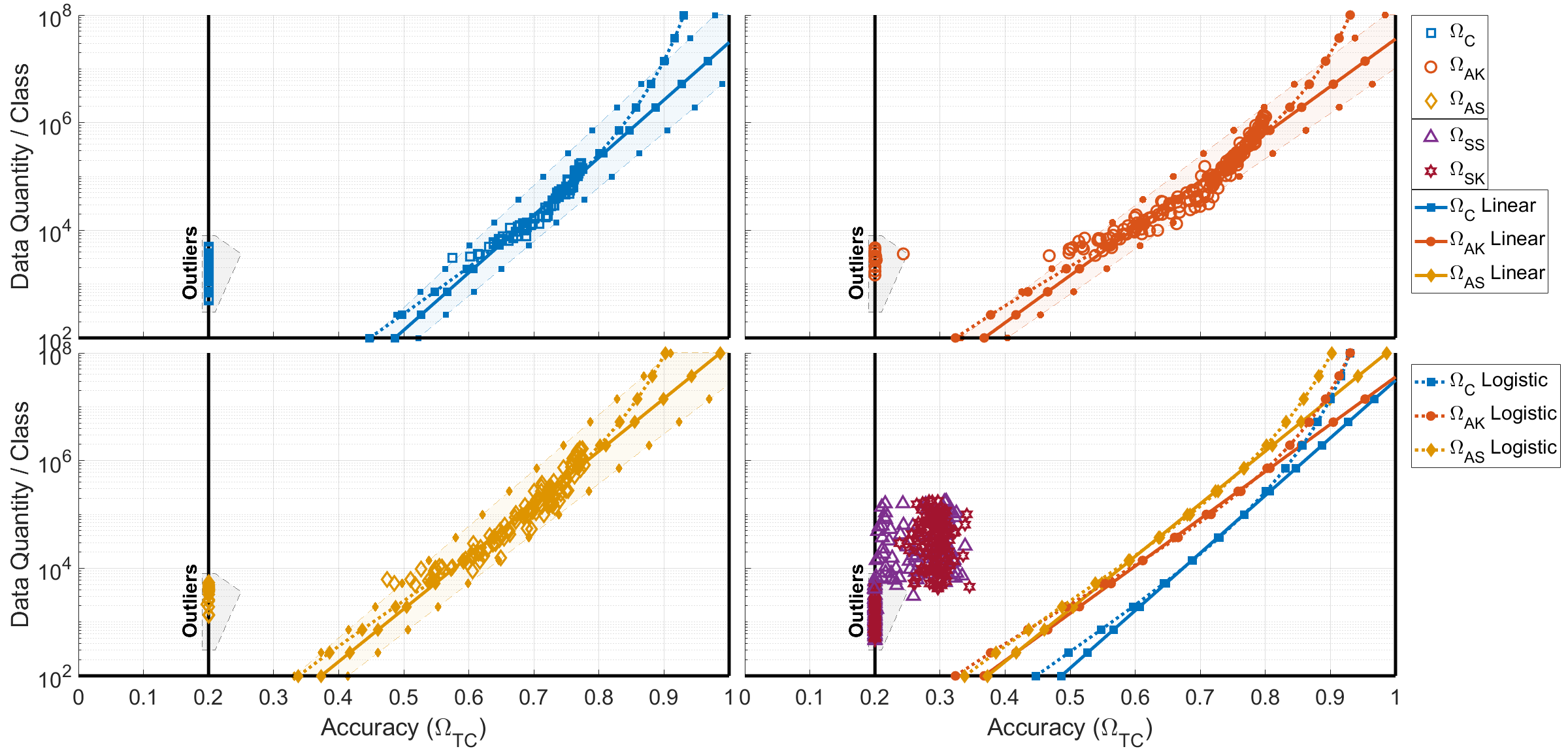}
    \caption{Performance of models trained using the five datasets on the $\Phi_{5}$ waveform space. The solid vertical line represents a network that is performing as well as a random guess. Solid lines with markers represent the trend in terms of examples per class that are needed to achieve a given accuracy on the capture test data, while the shaded regions between dashed lines with the matching markers indicate a 95\% confidence region for that trend. Synthetic datasets are omitted from the trend analysis as no significant trend was observed from these datasets. Trends are derived with outliers removed. The further right the trend line, the higher the quality of data. Additionally, a dotted line is fit to the data using a logistic regression with the assumption that 100\% accuracy is possible given the asymptotic curve perceived in the data and the impossibility of performance greater than 100\% accuracy. The slopes of the linear fits decreases as the problem is more complex than in Figure \ref{fig:perf_capt_qty_03}.}
    \label{fig:perf_capt_qty_05}
    
    \cleardoublepage
    \cleardoublepage
    \cleardoublepage
\end{figure*}

\begin{figure*}[t]
    \includegraphics[width=\textwidth,trim=0 0 0 0, clip]{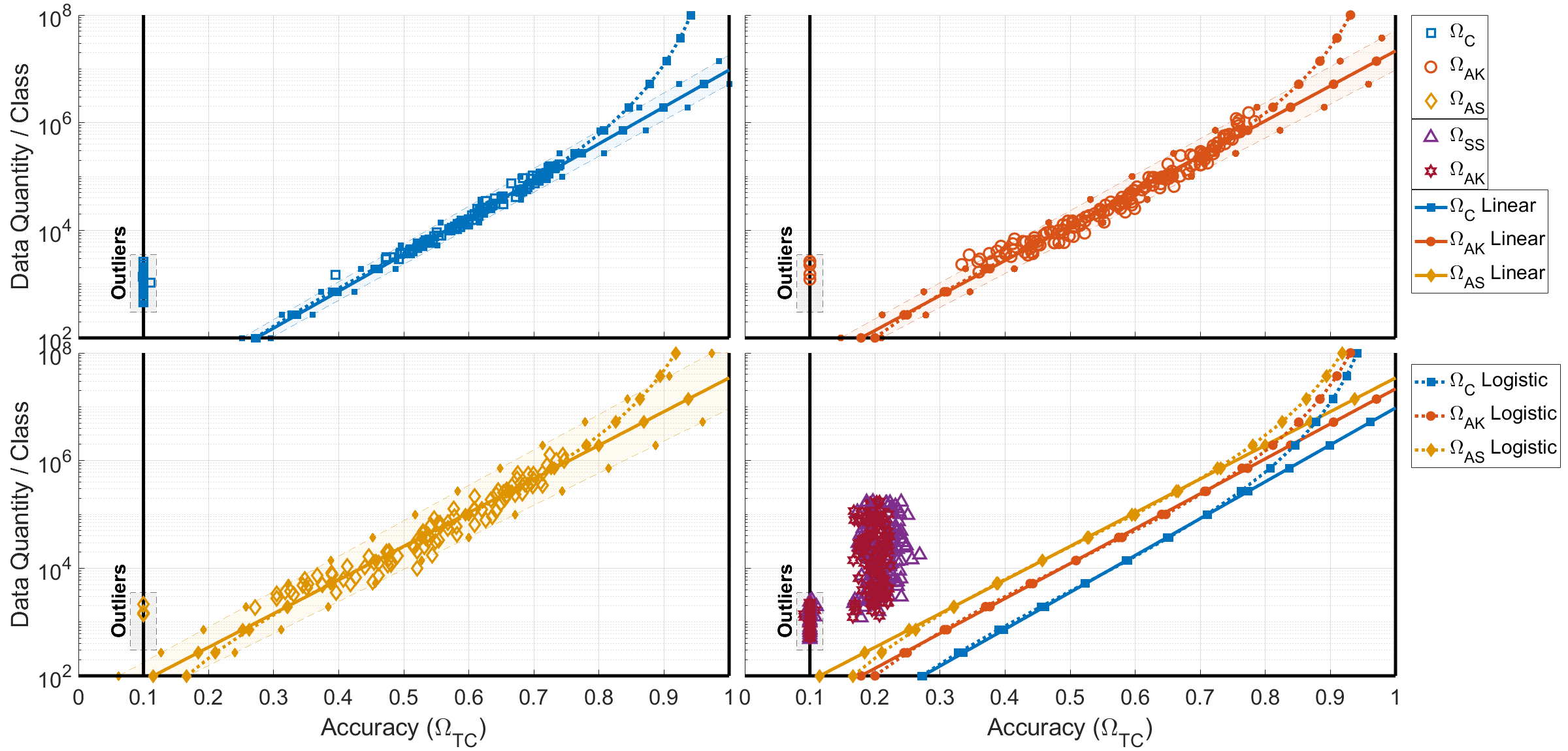}
    \caption{Performance of models trained using the five datasets on the $\Phi_{10}$ waveform space. The solid vertical line represents a network that is performing as well as a random guess. Solid lines with markers represent the trend in terms of examples per class that are needed to achieve a given accuracy on the capture test data, while the shaded regions between dashed lines with the matching markers indicate a 95\% confidence region for that trend. Synthetic datasets are omitted from the trend analysis as no significant trend was observed from these datasets. Trends are derived with outliers removed. The further right the trend line, the higher the quality of data. Additionally, a dotted line is fit to the data using a logistic regression with the assumption that 100\% accuracy is possible given the asymptotic curve perceived in the data and the impossibility of performance greater than 100\% accuracy. The slopes of the linear fits decrease as the problem is more complex than in Figure \ref{fig:perf_capt_qty_05}.}
    \label{fig:perf_capt_qty_10}
\end{figure*}

Figures \ref{fig:perf_capt_qty_03}-\ref{fig:perf_capt_qty_10} show the relationship between the achieved performance on $\Omega_{TC}$ of each individually trained network on the x-axis, with the y-axis corresponding to the total number of uniquely stored observations available during the training of the network.
By looking at the relation between accuracy achieved and total data per class used during the training, two important pieces of information can be extracted.
First, the trend lines further to the right for a given total quantity exhibit a higher quality within the data, because a better performance is achievable.
Using this, the quality of data decreases in the following order of the capture datasets across all examined waveform spaces: $\Omega_{C}$, $\Omega_{AK}$, $\Omega_{AS}$.
Second, assuming the log-linear trend holds, as shown in Table \ref{tab:fits}, without an asymptotic bound on accuracy (an asymptotic bound should be expected, and is given with the dotted lines), a forecast can be made on just how much data of each type is required in order to achieve ideal performance, and these quantity values are shown in Table \ref{tab:quantity}, with the total continuous capture time required to perform such a capture as has been done for this dataset for each waveform space is given in Table \ref{tab:timelapse} in terms of days.
As the trends are not consistent across all waveform spaces, also plotted is the 95\% confidence region around those linear trends in shaded regions bounded by dashed lines with the same marker.
However, assuming that the trends are consistent given enough observations, the general results align well with intuition in that data captured directly from the test environment is of highest quality and needs the least number of observations to achieve a target performance for the given model architecture and training routine.
Second to the captured data, augmentation of data to match the nuisance parameter distributions from the test environment provides the next highest quality data for training followed by naive augmentation that doesn't consider the full nuisance parameter distributions.
Coming in last, by many orders of magnitude, is synthetic data that only considers detection imperfections while simulating the waveform spaces.


Instead of using log-linear trends, using a log-logistic parametric fits parameters shown in Table \ref{tab:fits_log} results in the observation quantities in Table \ref{tab:quantity_log} in order to achieve a 95\% accuracy in the problem space, given that 100\% accuracy would require an infinite quantity of data.
This results in the more likely capture durations shown in Table \ref{tab:timelapse_log}.
These results show that for the smallest waveform space an order of magnitude less capture duration can be performed for giving up the 5\% accuracy, whereas for the larger waveform spaces roughly 20 times longer capture duration will be needed even while giving up that 5\% performance.
One more note is that the logistic regression assumes that 100\% accuracy is an asymptotically achievable feat, while it is more likely that given the model and training style that the peak performance achievable would still be less than 100\%.

\begin{table}
\small\sf\centering
\caption{Quantity of examples per class needed to achieve 100\% accuracy for each dataset source and waveform space. Extrapolated from linear trends in Figures \ref{fig:perf_capt_qty_03}-\ref{fig:perf_capt_qty_10}. Assuming no asymptotic limit.}\label{tab:quantity}
\begin{tabular}{c c c c}
\toprule
\multirow{2}{*}{Dataset} & \multicolumn{3}{c}{Waveform Space}\\
 & $\Phi_{3}$ & $\Phi_{5}$ & $\Phi_{10}$\\
\midrule
$\Omega_{C}$ & $58.9\mathrm{e}{6}$ & $31.0\mathrm{e}{6}$ & $9.5\mathrm{e}{6}$\\
\midrule
$\Omega_{AK}$ & $17.2\mathrm{e}{6}$ & $35.8\mathrm{e}{6}$ & $21.5\mathrm{e}{6}$\\
\midrule
$\Omega_{AS}$ & $24.3\mathrm{e}{6}$ & $135.5\mathrm{e}{6}$ & $34.3\mathrm{e}{6}$\\
\bottomrule
\end{tabular}
\end{table}
\begin{table}
\small\sf\centering
\caption{Quantifying the duration of a continuous capture, with no down time needed, in order to capture all data required to fulfill the $\Omega_{C}$ requirement for each waveform space assuming a 40kHz sampling rate of a 5kHz baud rate signal in Days. Assuming no asymptotic limit.}\label{tab:timelapse}
\begin{tabular}{c c c c}
\toprule
\multirow{2}{*}{Dataset} & \multicolumn{3}{c}{Waveform Space}\\
 & $\Phi_{3}$ & $\Phi_{5}$ & $\Phi_{10}$\\
\midrule
$\Omega_{C}$ & $104.7$ & $91.7$ & $56.4$\\
\bottomrule 
\end{tabular}
\end{table}

\begin{table}
\small\sf\centering
\caption{Log-Logistic fits, $\text{qty}=10^{-\left(\log\left(\frac{1-\alpha}{\alpha}\right)/p_1 - p_2\right)}$, for data presented in Figures \ref{fig:perf_capt_qty_03}-\ref{fig:perf_capt_qty_10}.}\label{tab:fits_log}
\begin{tabular}{c c c}
\toprule
\multirow{2}{*}{Dataset} & \multicolumn{2}{c}{Waveform Space $(p_1,p_2)$}\\
    & $\Phi_{3}$ & $\Phi_{5}$ \\
\midrule
$\Omega_{C}$ &  $0.3452,-1.705$ & $0.4674,2.449$\\
\midrule
$\Omega_{AK}$ & $0.5275,1.015$ & $0.5548,3.328$\\
\midrule
$\Omega_{AS}$ & $0.4944,1.094$ & $0.4821,3.393$\\
\midrule
& \multicolumn{2}{c}{$\Phi_{10}$}\\
\midrule
$\Omega_{C}$ & \multicolumn{2}{c}{$0.6274,3.573$}\\
\midrule
$\Omega_{AK}$ & \multicolumn{2}{c}{$0.6641,4.087$}\\
\midrule
$\Omega_{AS}$ & \multicolumn{2}{c}{$0.6714,4.399$}\\
\bottomrule
\end{tabular}
\end{table}

\begin{table}
\small\sf\centering
\caption{Quantity of examples per class needed to achieve 95\% accuracy for each dataset source and waveform space. Extrapolated from logistic fit in Figures \ref{fig:perf_capt_qty_03}-\ref{fig:perf_capt_qty_10}.}\label{tab:quantity_log}
\begin{tabular}{c c c c}
\toprule
\multirow{2}{*}{Dataset} & \multicolumn{3}{c}{Waveform Space}\\
    & $\Phi_{3}$ & $\Phi_{5}$ & $\Phi_{10}$\\
\midrule
$\Omega_{C}$ & $6.7\mathrm{e}{6}$ & $560.4\mathrm{e}{6}$ & $184.5\mathrm{e}{6}$\\
\midrule
$\Omega_{AK}$ & $3.9\mathrm{e}{6}$ & $431.7\mathrm{e}{6}$ & $331.6\mathrm{e}{6}$\\
\midrule
$\Omega_{AS}$ & $11.2\mathrm{e}{6}$ & $3169.8\mathrm{e}{6}$ & $609.1\mathrm{e}{6}$\\
\bottomrule
\end{tabular}
\end{table}

\begin{table}
\small\sf\centering
\caption{Quantifying the duration of a continuous capture, with no down time needed, in order to capture all data required to fulfill the 95\% accuracy requirement using dataset $\Omega_{C}$ for each waveform space assuming a 40kHz sampling rate of a 5kHz baud rate signal in Days.}\label{tab:timelapse_log}
\begin{tabular}{c c c c}
\toprule
\multirow{2}{*}{Dataset} & \multicolumn{3}{c}{Waveform Space}\\
    & $\Phi_{3}$ & $\Phi_{5}$ & $\Phi_{10}$\\
\midrule
$\Omega_{C}$ & $11.9$ & $1660.7$ & $1093.4$\\
\bottomrule
\end{tabular}
\end{table}

One question that naturally follows this quality comparison of the datasets is then if augmented data is of lower quality, then why not just focus on getting more captured data?
The primary reason for relying on augmentation is cost, both in terms of time and money.
In terms of time, the capture dataset was collected over a 4-month window in 2018, while the augmented datasets were generated over the course of 2-4 days each and contain an order of magnitude more observations per dataset.
For full comparison, the synthetic datasets were generated over the course of 7 days for each dataset and are of the same order of magnitude as the captured data. One contributing factor for the increased generation time of the synthetic data was the design decision of extracting only one observation per execution of GNU Radio flowgraph, rather than extracting many observations from a single execution, which was done to decrease any dependence between observations within the dataset.
The second cost is the monetary expenditures for procuring the transceivers, and making them robust enough to last 4 months of continuous use, paying for the power and space needed to make the transmissions, and the personnel for setting up and maintaining the capture.
Determining the value of data is beyond the scope of this work.

\begin{figure*}[t]
    \includegraphics[width=\textwidth,trim=0 0 0 0, clip]{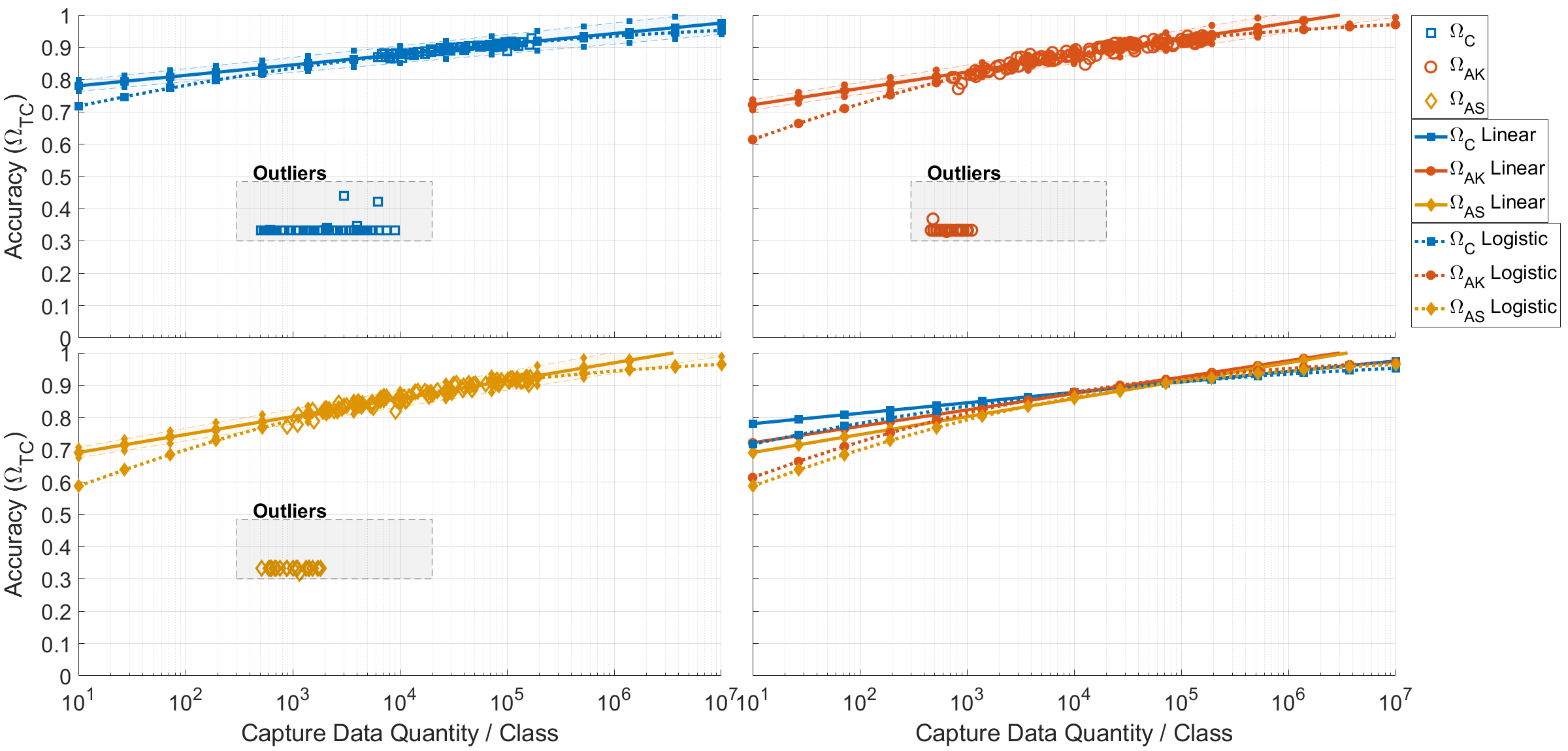}
    \caption{Performance of models trained using the three capture datasets on the $\Phi_{3}$ waveform space. Solid lines represent the trend in terms of examples per class that are needed from $\Omega_{C}$ to achieve a given accuracy on the capture test data with or without any augmentation, while the shaded regions between dashed lines with the matching markers indicate a 95\% confidence region for that trend. Trends are derived with outliers removed. The higher the trend line, the higher the quality of the dataset. The dotted lines represent a log-logistic regression to account for the asymptotic curvature observed in the results. Under both sets of regression, the models using dataset $\Omega_{AK}$ exhibit a higher quality data with increasing data quantity.}
    \label{fig:perf_capt_03}
\end{figure*}

\begin{figure*}[t]
    \includegraphics[width=\textwidth,trim=0 0 0 0, clip]{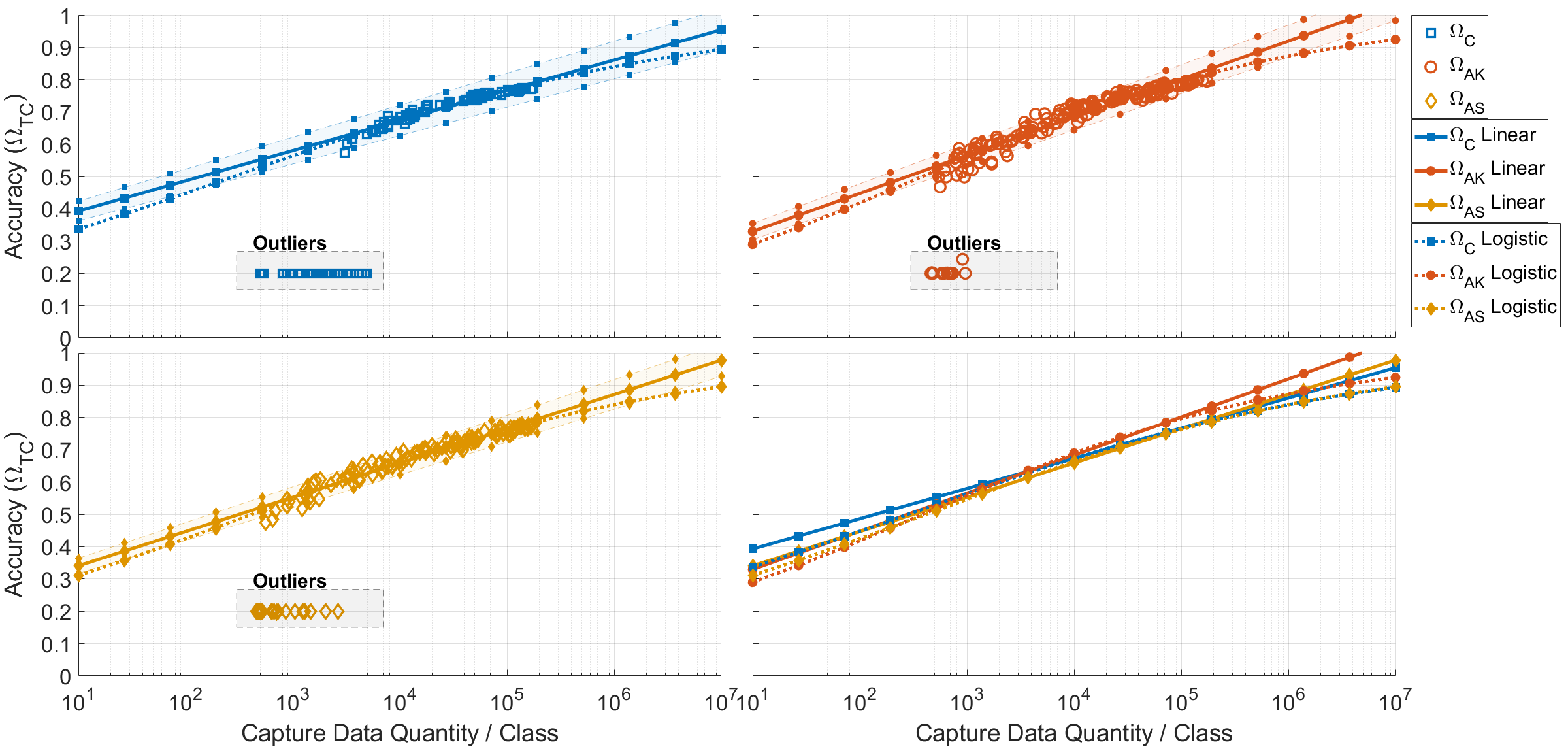}
    \caption{Performance of models trained using the three capture datasets on the $\Phi_{5}$ waveform space. Solid lines represent the trend in terms of examples per class that are needed from $\Omega_{C}$ to achieve a given accuracy on the capture test data with or without any augmentation, while the shaded regions between dashed lines with the matching markers indicate a 95\% confidence region for that trend. Trends are derived with outliers removed. The higher the trend line, the higher the quality of the dataset. The dotted lines represent a log-logistic regression to account for the asymptotic curvature observed in the results. Under both sets of regression, the models using dataset $\Omega_{AK}$ exhibit a higher quality data with increasing data quantity.}
    \label{fig:perf_capt_05}
\end{figure*}

\begin{figure*}[t]
    \includegraphics[width=\textwidth,trim=0 0 0 0, clip]{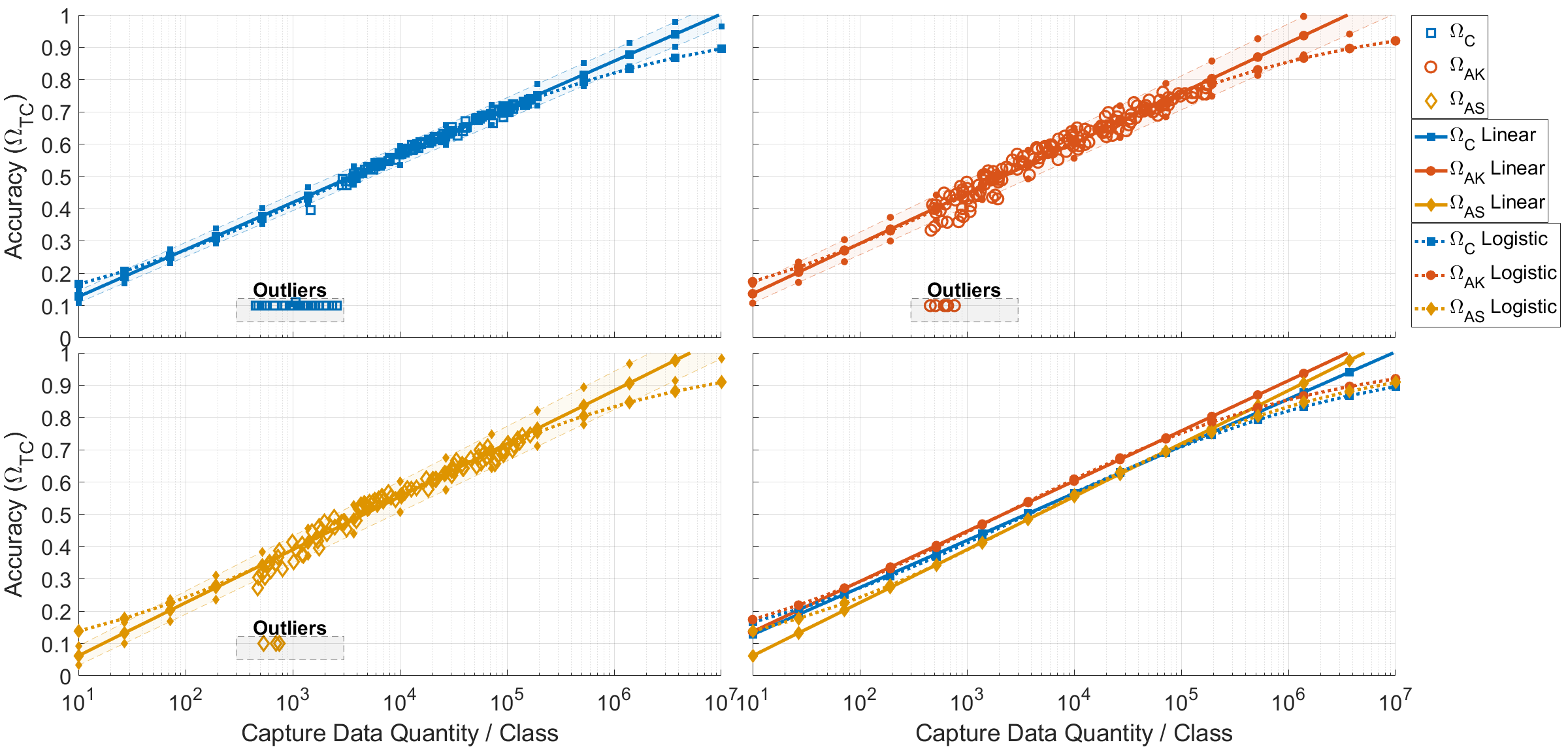}
    \caption{Performance of models trained using the three capture datasets on the $\Phi_{10}$ waveform space. Solid lines represent the trend in terms of examples per class that are needed from $\Omega_{C}$ to achieve a given accuracy on the capture test data with or without any augmentation, while the shaded regions between dashed lines with the matching markers indicate a 95\% confidence region for that trend. Trends are derived with outliers removed. The higher the trend line, the higher the quality of the dataset. The dotted lines represent a log-logistic regression to account for the asymptotic curvature observed in the results. In contrast to Figures \ref{fig:perf_capt_03} and \ref{fig:perf_capt_05} the full observed trends from networks $\Omega_{AK}$ exceed the trends when only $\Omega_{C}$ is used indicating the benefit of augmentation as the complexity of the problem space increases.}
    \label{fig:perf_capt_10}
\end{figure*}

So far the results have been shown in total number of observations used, but there is one more important way to look at the augmentation performance, and that there must be some foundation of capture data from which to augment.
Figures \ref{fig:perf_capt_03}-\ref{fig:perf_capt_10} shuffle the results of the capture and augmented datasets to show the accuracy achieved on $\Omega_{TC}$ as a function of the capture data quantity that went into each model's training.
This means that for a given value on the x-axis, all data points required the same number of capture observations per waveform class in order to achieve the performance shown.
What these figures do not show is the augmentation factor used by each augmented network result.
In this work, the augmentation factor is upper bounded by 10 due to the choice of having the augmentations performed prior to training the network and the storage constraints of the servers used, rather than augmentation performed online during training that would be one-offs unique to each augmented network.
From Figures \ref{fig:perf_capt_03}-\ref{fig:perf_capt_10}, two more beneficial aspects of augmented data can be observed.
The first beneficial aspect of augmentation allows for network convergence when the number of capture observations is not substantial enough to converge on their own.
This is tremendously beneficial when planning for a capture event and determining how long the event must be in order to achieve a desired performance level by establishing the trends like what was done in Table \ref{tab:quantity}, but performed in an order of magnitude smaller time window as an exploratory capture event.
The second benefit is seen when there are only a set number of observations available within the capture dataset, and knowledge about the degradation due to the detection algorithm, which is known, as under these conditions the accuracy of the networks trained with augmentation exceed those of the networks with only capture data alone.
From these results, while remembering that the augmentation used in this work is a static augmented dataset with a bounded number of augmentations set to 10; the full effect of augmentation and how performance changes with dynamic, large augmentation factors ($>$10) and as to whether there are any diminishing returns as the augmentation factor increases is outside the scope of this work and is an area for future work.


\subsection{Degradation Distribution Effect on Augmentation}\label{sect:degradation}
The last insight offered by Figures \ref{fig:perf_capt_qty_03}-\ref{fig:perf_capt_10} addresses the third question of whether knowing the distribution of the degradation is beneficial for augmentation.
Through contrasting the performance of $\Omega_{AK}$, where augmentation draws from the nuisance parameter KDE, with that of $\Omega_{AS}$, where augmentation is performed on an assumed subset of the application space, the quality of the augmented data can be examined.

For each of these figures, in the bottom right plot the log-linear and log-logistic trends for datasets $\Omega_{C}$, $\Omega_{AK}$, and $\Omega_{AS}$ are overlaid.
As the difficulty of the problem space increases, with regard to the plots for both the total available data (Figures \ref{fig:perf_capt_qty_03}-\ref{fig:perf_capt_qty_10}) and looking at the performance for a fixed quantity of captured data (Figures \ref{fig:perf_capt_03}-\ref{fig:perf_capt_10}), the trends derived from models trained with data using $\Omega_{AK}$ consistently outperform the trends derived from models using $\Omega_{AS}$.
Examining the trends within Figure \ref{fig:perf_capt_10} for the most complex waveform space considered in this work, $\Phi_{10}$, the trends for performance using the $\Omega_{AK}$ dataset are projected to be greater than the trends of the captured data alone for the full span of considered quantity.
While projections using the $\Omega_{AS}$ dataset do not exceed those of from the captured data alone until around the order 72,000 observations per class, and even then only just exceed the captured data alone by roughly 1.5\% in the log-logistic case.
Across all waveform spaces when considering the amount of capture data needed as shown in Figures \ref{fig:perf_capt_03}-\ref{fig:perf_capt_10}, the performance gains seen from using $\Omega_{AS}$ and contrasting that to the performance of the capture data on its own results in worse performance as long as there is enough data available to result in a convergent network while using the capture data.
Now when the only goal is peak performance, the expertise is lacking to evaluate the available data, and sufficient capture data is available, then the naive augmentation will help achieve a better result than the capture data alone, but with practical constraints of time and money this approach might prove to be impractical.

One further consideration not addressed in this work is the difference in the transmitter and modulation specific correlations.
To understand this point of contrast, all observations were between two distinct software radios capable of transmitting every waveform of interest, and therefore the data presented here does not take into consideration any correlations between the transmitting device and imperfections observed.
One such unique consideration is that a Global System for Mobile Communications (GSM) waveform from a base station should have a distinct set of degradations that would help improve the identification of the Gaussian Minimum Shift Keying (GMSK) waveform over a hand held two-way radio's Frequency Shift Keying (FSK) for a given environment.
Applying degradation from one type of transmitter (two-way hand held) to another modulation (GMSK) could end up hurting the performance of the overall system.
Investigating the effect of COTS systems and their distinct characteristics in contrast to more versatile software-defined radios should be another avenue investigated for enhancing the performance of a system that must perform with a wider range of transceivers present.

These observations suggest that care should be taken when creating augmentation routines such that the distributions on the nuisance parameters are considered during the augmentation in order to achieve peak performance for a given number of capture observations.
Additionally, naive methods of augmentation utilizing GANs should be cautious of developing one network capable of augmenting \emph{any} waveform, though developing a network \emph{per} waveform as was done in Davaslioglu et al. still might be a viable alternative to the domain knowledge approach in this work.\cite{adv_rfml:Davaslioglu2018a}

\subsection{Results Summary}\label{sect:resultssummary}
From this work, the results show that synthetic data that only considers degradations inherent in detector and isolation algorithms such as SNR, FO, and SRM, do not provide high enough fidelity data when evaluating on real-world datasets with dynamic channels.
This was the case for using the assumed subset and KDE parameter draws for synthesizing the degradation, but this result should not be taken as synthetic has no value, rather just not enough value as prepared and used in this work because an increase in performance was observed in $\Phi_{3}$ and $\Phi_{5}$ waveform spaces with using $\Omega_{SK}$ over $\Omega_{SS}$ when only a small portion of the overall environment was considered.
A stark contrast to the synthetic data was found in the value of augmentation of the dataset while only considering SNR, FO, and SRM.
If the augmentation was drawn from the captured distributions observed, a mixture of increased accuracy and generalization was observed in the network's final performance on the two test sets, $\Omega_{TC}$ and $\Omega_{TS}$.
However, when augmentation was focused on a subset of the distribution regions for the same degradation types, the end resulting network saw a decrease in performance for both test sets.
This shows that augmentation used properly can offer gains when taking into consideration the full range and distribution of what is expected to be observed, but naive usage could lead to performance loss instead.
Both augmentation sources did have one benefit over raw capture when there was not enough of the capture data for the network to converge: in this case, both naive and KDE based augmentation can be used to provide a rough estimate how much real-world captured data would be needed to achieve a desired performance, ignoring any asymptotic limit.
These results are drawn from the variation of the data used in this work, and the constant choices for the model architecture and training routine.
In order for these results to be held true across the RFML problem space, additional experiments are required with different model architectures and training regimes with similar outcomes confirming these results.

\section{Future Work}\label{sect:future}
New questions or efforts that can build upon these results include:
\begin{itemize}
    \item Expand the observation space utilized within this work to consider other training routines and networks to reinforce or contradict the results of this work. (Section \ref{sect:experiment})
    \item Integration of propagation path operational requirements into dataset generation (Section \ref{sect:thecapturedataset})
    \item Testing the predicted quantities needed to reach 100\% accuracy made by this paper with larger datasets, or identifying an asymptotic limit (Section \ref{sect:valueaugment})
    \item Quantifying the relationship of the augmentation factor to performance, conditionally on the amount of capture data available (Section \ref{sect:valueaugment})
    \item Quantifying cost of training datasets for a given RFML application (Section \ref{sect:degradation}) 
    \item Quantify the ability of capture data to be augmented for a different type and quantity of propagation paths (Section \ref{sect:degradation})
    \item Incorporate the emulation of RF signal streams through channel environments that cannot be practically tested, like atmospheric scintillation \cite{Sward} (Section \ref{sect:degradation})
    \item Examine the viability of generative network to perform augmentation without domain knowledge explicitly given (Section \ref{sect:resultssummary})
    \item Generalization of parametric training data quantification to other RFML applications (Section \ref{sect:resultssummary})
\end{itemize}

\section{Conclusions}\label{sect:conclusion}
Three questions are examined and addressed within this work.
First, the results show that only considering the nuisance parameters from the detection and isolation algorithms do not provide enough to bridge the synthetic and real-world divide, and suggest that in order to bring synthetic data generation to a higher fidelity, the propagation path from DAC to ADC must be further investigated and modeled.
The second question finds that while the overall quality of augmented observations are less than that of uniquely captured observations, the associated cost in terms of time and money is significantly lower with augmented data, suggesting a cost analysis can be performed to strike a balance between the two.
Additionally, the use of augmentation when there is not enough data, or capturing more data is not a feasible option, allows for an increase in performance over just the captured data on its own, especially when the capture dataset is insufficient to allow for the network to converge during training.
The final take away is shown in the results to conclusively align with knowledge of the distributions being used while performing augmentation are consistently better than naively augmenting data with an assumed near-set parameter space.
The work establishes a methodology to make a prediction for the quantity of the data needed, under all cases examined, for the number of observations needed to reach 100\% accuracy in the classification problem for the provided dataset, not accounting for any asymptotic limit existing prior to reaching 100\% accuracy, as well as for determining the number of observations needed to reach 95\% accuracy with logistic regression when taking into consideration the asymptotic limit of the problem space.

\begin{acks}
The authors wish to thank CACI for their sponsorship of the original OTA dataset collection.
\end{acks}

\bibliographystyle{SageV}

\begin{thebibliography}{10}
    \providecommand{\url}[1]{\texttt{#1}}
    \providecommand{\urlprefix}{URL }
    \expandafter\ifx\csname urlstyle\endcsname\relax
      \providecommand{\doi}[1]{DOI:\discretionary{}{}{}#1}\else
      \providecommand{\doi}{DOI:\discretionary{}{}{}\begingroup
      \urlstyle{rm}\Url}\fi
    \providecommand{\eprint}[2][]{\url{#2}}
    
    \bibitem{jesmo}
    {Joint Chiefs of Staff, Joint Publication 3-85: ``Joint Electromagnetic
      Spectrum Operations,'' 22 May 2020}.
    
    \bibitem{5g}
    {Morocho-Cayamcela} ME, {Lee} H and {Lim} W.
    \newblock {Machine Learning for {5G/B5G} Mobile and Wireless Communications:
      Potential, Limitations, and Future Directions}.
    \newblock \emph{IEEE Access} 2019; 7: 137184--137206.
    \newblock \doi{10.1109/ACCESS.2019.2942390}.
    
    \bibitem{signaleye}
    {{SignalEye} {AI} Software for Automated Signal Classification - {General
      Dynamics}}.
    \newblock
      \urlprefix\url{https://gdmissionsystems.com/products/electronic-warfare/signaleye}.
    
    \bibitem{wong-sei1}
    {Wong} LJ, {Headley} WC, {Andrews} S et~al.
    \newblock {Clustering Learned {CNN} Features from Raw {I/Q} Data for Emitter
      Identification}.
    \newblock In \emph{MILCOM 2018 - 2018 IEEE Military Communications Conference
      (MILCOM)}. pp. 26--33.
    \newblock \doi{10.1109/MILCOM.2018.8599847}.
    
    \bibitem{wong-sei2}
    {Wong} LJ, {Headley} WC and {Michaels} AJ.
    \newblock {Specific Emitter Identification Using Convolutional Neural
      Network-Based {IQ} Imbalance Estimators}.
    \newblock \emph{IEEE Access} 2019; 7: 33544--33555.
    
    \bibitem{sankhe2019oracle}
    Sankhe K, Belgiovine M, Zhou F et~al.
    \newblock {ORACLE: Optimized Radio clAssification through Convolutional neuraL
      nEtworks}.
    \newblock In \emph{IEEE INFOCOM 2019-IEEE Conference on Computer
      Communications}. IEEE, pp. 370--378.
    
    \bibitem{Goodfellow-et-al-2016}
    Goodfellow I, Bengio Y and Courville A.
    \newblock \emph{{Deep Learning}}.
    \newblock MIT Press, 2016.
    \newblock \url{http://www.deeplearningbook.org}.
    
    \bibitem{Nandi_1997}
    Nandi A and Azzouz E.
    \newblock Modulation recognition using artificial neural networks.
    \newblock \emph{Signal Processing} 1997; 56(2): 165 -- 175.
    \newblock \doi{https://doi.org/10.1016/S0165-1684(96)00165-X}.
    \newblock
      \urlprefix\url{http://www.sciencedirect.com/science/article/pii/S016516849600165X}.
    
    \bibitem{Kim_2003}
    {Namjin Kim}, {Kehtarnavaz} N, {Yeary} MB et~al.
    \newblock {{DSP}-based hierarchical neural network modulation signal
      classification}.
    \newblock \emph{IEEE Transactions on Neural Networks} 2003; 14(5): 1065--1071.
    
    \bibitem{Fehske_2005}
    {Fehske} A, {Gaeddert} J and {Reed} JH.
    \newblock {A new approach to signal classification using spectral correlation
      and neural networks}.
    \newblock In \emph{First IEEE International Symposium on New Frontiers in
      Dynamic Spectrum Access Networks, 2005. DySPAN 2005.} pp. 144--150.
    \newblock \doi{10.1109/DYSPAN.2005.1542629}.
    
    \bibitem{Mody_2007}
    {Mody} AN, {Blatt} SR, {Mills} DG et~al.
    \newblock {Recent advances in cognitive communications}.
    \newblock \emph{IEEE Communications Magazine} 2007; 45(10): 54--61.
    \newblock \doi{10.1109/MCOM.2007.4342823}.
    
    \bibitem{Ge_2008}
    {Ge} F, {Chen} Q, {Wang} Y et~al.
    \newblock {Cognitive Radio: From Spectrum Sharing to Adaptive Learning and
      Reconfiguration}.
    \newblock In \emph{2008 IEEE Aerospace Conference}. pp. 1--10.
    \newblock \doi{10.1109/AERO.2008.4526372}.
    
    \bibitem{Bixio_2009}
    {Bixio} L, {Ottonello} M, {Sallam} H et~al.
    \newblock {Signal classification based on spectral redundancy and neural
      network ensembles}.
    \newblock In \emph{2009 4th International Conference on Cognitive Radio
      Oriented Wireless Networks and Communications}. pp. 1--6.
    \newblock \doi{10.1109/CROWNCOM.2009.5189036}.
    
    \bibitem{adv_rfml:Clancy2009a}
    {Clancy} TC and {Khawar} A.
    \newblock {Security threats to signal classifiers using self-organizing maps}.
    \newblock In \emph{2009 4th International Conference on Cognitive Radio
      Oriented Wireless Networks and Communications}. pp. 1--6.
    \newblock \doi{10.1109/CROWNCOM.2009.5189050}.
    
    \bibitem{Ramon_2009}
    {Ramón} MM, {Atwood} T, {Barbin} S et~al.
    \newblock {Signal classification with an {SVM-FFT} approach for feature
      extraction in cognitive radio}.
    \newblock In \emph{2009 SBMO/IEEE MTT-S International Microwave and
      Optoelectronics Conference (IMOC)}. pp. 286--289.
    \newblock \doi{10.1109/IMOC.2009.5427579}.
    
    \bibitem{Popoola_2011}
    {Popoola} JJ and v~{Olst} R.
    \newblock {A Novel Modulation-Sensing Method}.
    \newblock \emph{IEEE Vehicular Technology Magazine} 2011; 6(3): 60--69.
    \newblock \doi{10.1109/MVT.2011.941893}.
    
    \bibitem{Kang_2011}
    {Shan Kang}, {Naiwen Chen}, {Mi Yan} et~al.
    \newblock {Detecting identity-spoof attack based on BP network in cognitive
      radio network}.
    \newblock In \emph{Proceedings of 2011 Cross Strait Quad-Regional Radio Science
      and Wireless Technology Conference}, volume~2. pp. 1603--1606.
    \newblock \doi{10.1109/CSQRWC.2011.6037280}.
    
    \bibitem{Pu_2011b}
    {Pu} D and {Wyglinski} AM.
    \newblock {Primary user emulation detection using frequency domain action
      recognition}.
    \newblock In \emph{Proceedings of 2011 IEEE Pacific Rim Conference on
      Communications, Computers and Signal Processing}. pp. 791--796.
    \newblock \doi{10.1109/PACRIM.2011.6032995}.
    
    \bibitem{He_2011}
    {He} F, {Xu} X, {Zhou} L et~al.
    \newblock {A learning based cognitive radio receiver}.
    \newblock In \emph{2011 - MILCOM 2011 Military Communications Conference}. pp.
      7--12.
    \newblock \doi{10.1109/MILCOM.2011.6127675}.
    
    \bibitem{Abedlreheem_2012}
    {Abdelreheem} MMT and {Helmi} MO.
    \newblock {Digital Modulation Classification through time and frequency domain
      features using Neural Networks}.
    \newblock In \emph{2012 IX International Symposium on Telecommunications
      (BIHTEL)}. pp. 1--5.
    \newblock \doi{10.1109/BIHTEL.2012.6412073}.
    
    \bibitem{Li_2012}
    {Li} S, {Wang} X and {Wang} J.
    \newblock {Manifold learning-based automatic signal identification in cognitive
      radio networks}.
    \newblock \emph{IET Communications} 2012; 6(8): 955--963.
    \newblock \doi{10.1049/iet-com.2010.0590}.
    
    \bibitem{Thilina_2013}
    {Thilina} KM, {Choi} KW, {Saquib} N et~al.
    \newblock {Machine Learning Techniques for Cooperative Spectrum Sensing in
      Cognitive Radio Networks}.
    \newblock \emph{IEEE Journal on Selected Areas in Communications} 2013; 31(11):
      2209--2221.
    \newblock \doi{10.1109/JSAC.2013.131120}.
    
    \bibitem{Popoola_2013}
    Popoola JJ and van Olst R.
    \newblock {The performance evaluation of a spectrum sensing implementation
      using an automatic modulation classification detection method with a
      Universal Software Radio Peripheral}.
    \newblock \emph{Expert Systems with Applications} 2013; 40(6): 2165 -- 2173.
    \newblock \doi{https://doi.org/10.1016/j.eswa.2012.10.047}.
    \newblock
      \urlprefix\url{http://www.sciencedirect.com/science/article/pii/S0957417412011712}.
    
    \bibitem{Kim_2013}
    {Kim} S and {Giannakis} GB.
    \newblock {Dynamic learning for cognitive radio sensing}.
    \newblock In \emph{2013 5th IEEE International Workshop on Computational
      Advances in Multi-Sensor Adaptive Processing (CAMSAP)}. pp. 388--391.
    \newblock \doi{10.1109/CAMSAP.2013.6714089}.
    
    \bibitem{Tsakmalis_2014}
    {Tsakmalis} A, {Chatzinotas} S and {Ottersten} B.
    \newblock {Automatic Modulation Classification for adaptive Power Control in
      cognitive satellite communications}.
    \newblock In \emph{2014 7th Advanced Satellite Multimedia Systems Conference
      and the 13th Signal Processing for Space Communications Workshop
      (ASMS/SPSC)}. pp. 234--240.
    \newblock \doi{10.1109/ASMS-SPSC.2014.6934549}.
    
    \bibitem{Chen_2015}
    {Chen} T, {Liu} J, {Xiao} L et~al.
    \newblock {Anti-jamming transmissions with learning in heterogenous cognitive
      radio networks}.
    \newblock In \emph{2015 IEEE Wireless Communications and Networking Conference
      Workshops (WCNCW)}. pp. 293--298.
    \newblock \doi{10.1109/WCNCW.2015.7122570}.
    
    \bibitem{Mendis_2016}
    {Mendis} GJ, {Wei} J and {Madanayake} A.
    \newblock {Deep learning-based automated modulation classification for
      cognitive radio}.
    \newblock In \emph{2016 IEEE International Conference on Communication Systems
      (ICCS)}. pp. 1--6.
    \newblock \doi{10.1109/ICCS.2016.7833571}.
    
    \bibitem{Oshea_2016}
    {O'Shea} TJ, {Pemula} L, {Batra} D et~al.
    \newblock {Radio transformer networks: Attention models for learning to
      synchronize in wireless systems}.
    \newblock In \emph{2016 50th Asilomar Conference on Signals, Systems and
      Computers}. pp. 662--666.
    \newblock \doi{10.1109/ACSSC.2016.7869126}.
    
    \bibitem{Oshea_2016c}
    {O'Shea} TJ, {Hitefield} S and {Corgan} J.
    \newblock {End-to-end radio traffic sequence recognition with recurrent neural
      networks}.
    \newblock In \emph{2016 IEEE Global Conference on Signal and Information
      Processing (GlobalSIP)}. pp. 277--281.
    \newblock \doi{10.1109/GlobalSIP.2016.7905847}.
    
    \bibitem{oshea2016datagen}
    O'Shea T and West N.
    \newblock {Radio Machine Learning Dataset Generation with {GNU} Radio}.
    \newblock \emph{Proceedings of the GNU Radio Conference} 2016; 1(1).
    \newblock
      \urlprefix\url{https://pubs.gnuradio.org/index.php/grcon/article/view/11}.
    
    \bibitem{Oshea_2016d}
    {O'Shea} TJ, {Corgan} J and {Clancy} TC.
    \newblock {Convolutional Radio Modulation Recognition Networks}.
    \newblock \emph{arXiv e-prints} 2016; : arXiv:1602.04105\eprint{1602.04105}.
    
    \bibitem{west_2017}
    {West} NE, {Harwell} K and {McCall} B.
    \newblock {{DFT} signal detection and channelization with a deep neural network
      modulation classifier}.
    \newblock In \emph{2017 IEEE International Symposium on Dynamic Spectrum Access
      Networks (DySPAN)}. pp. 1--3.
    \newblock \doi{10.1109/DySPAN.2017.7920745}.
    
    \bibitem{west-amc2}
    {West} NE and {O'Shea} T.
    \newblock {Deep architectures for modulation recognition}.
    \newblock In \emph{2017 IEEE International Symposium on Dynamic Spectrum Access
      Networks (DySPAN)}. pp. 1--6.
    \newblock \doi{10.1109/DySPAN.2017.7920754}.
    
    \bibitem{Karra_2017}
    {Karra} K, {Kuzdeba} S and {Petersen} J.
    \newblock {Modulation recognition using hierarchical deep neural networks}.
    \newblock In \emph{2017 IEEE International Symposium on Dynamic Spectrum Access
      Networks (DySPAN)}. pp. 1--3.
    \newblock \doi{10.1109/DySPAN.2017.7920746}.
    
    \bibitem{oshea-detect2}
    {O'Shea} T, {Roy} T and {Clancy} TC.
    \newblock {Learning robust general radio signal detection using computer vision
      methods}.
    \newblock In \emph{2017 51st Asilomar Conference on Signals, Systems, and
      Computers}. pp. 829--832.
    
    \bibitem{Peng_2017}
    {Peng} S, {Jiang} H, {Wang} H et~al.
    \newblock {Modulation classification using convolutional Neural Network based
      deep learning model}.
    \newblock In \emph{2017 26th Wireless and Optical Communication Conference
      (WOCC)}. pp. 1--5.
    \newblock \doi{10.1109/WOCC.2017.7929000}.
    
    \bibitem{Reddy_2017}
    {Reddy} KPK, {Yeleswarapu} Y and {Darak} SJ.
    \newblock {Performance evaluation of cumulant feature based automatic
      modulation classifier on {USRP} testbed}.
    \newblock In \emph{2017 9th International Conference on Communication Systems
      and Networks (COMSNETS)}. pp. 393--394.
    \newblock \doi{10.1109/COMSNETS.2017.7945409}.
    
    \bibitem{Nawaz_2017}
    {Nawaz} T, {Marcenaro} L and {Regazzoni} CS.
    \newblock {Stealthy jammer detection algorithm for wide-band radios: A physical
      layer approach}.
    \newblock In \emph{2017 IEEE 13th International Conference on Wireless and
      Mobile Computing, Networking and Communications (WiMob)}. pp. 79--83.
    \newblock \doi{10.1109/WiMOB.2017.8115792}.
    
    \bibitem{Nawaz_2017b}
    Nawaz T, Marcenaro L and Regazzoni CS.
    \newblock {Cyclostationary-based jammer detection for wideband radios using
      compressed sensing and artificial neural network.}
    \newblock \emph{International Journal of Distributed Sensor Networks} 2017;
      13(12): 1.
    \newblock
      \urlprefix\url{http://login.ezproxy.lib.vt.edu/login?url=http://search.ebscohost.com/login.aspx?direct=true&db=iih&AN=127039926&scope=site}.
    
    \bibitem{Ambaw_2017}
    {Ambaw} AB, {Bari} M and {Doroslovački} M.
    \newblock {A case for stacked autoencoder based order recognition of
      continuous-phase {FSK}}.
    \newblock In \emph{2017 51st Annual Conference on Information Sciences and
      Systems (CISS)}. pp. 1--6.
    \newblock \doi{10.1109/CISS.2017.7926151}.
    
    \bibitem{Ali_2017}
    Ali A, Yangyu F and Liu S.
    \newblock Automatic modulation classification of digital modulation signals
      with stacked autoencoders.
    \newblock \emph{Digital Signal Processing} 2017; 71: 108 -- 116.
    \newblock \doi{https://doi.org/10.1016/j.dsp.2017.09.005}.
    \newblock
      \urlprefix\url{http://www.sciencedirect.com/science/article/pii/S1051200417302087}.
    
    \bibitem{Hong_2017}
    {Hong} D, {Zhang} Z and {Xu} X.
    \newblock {Automatic modulation classification using recurrent neural
      networks}.
    \newblock In \emph{2017 3rd IEEE International Conference on Computer and
      Communications (ICCC)}. pp. 695--700.
    \newblock \doi{10.1109/CompComm.2017.8322633}.
    
    \bibitem{Yelalwar_2018}
    {Yelalwar} RG and {Ravinder} Y.
    \newblock {Artificial Neural Network Based Approach for Spectrum Sensing in
      Cognitive Radio}.
    \newblock In \emph{2018 International Conference on Wireless Communications,
      Signal Processing and Networking (WiSPNET)}. pp. 1--5.
    \newblock \doi{10.1109/WiSPNET.2018.8538729}.
    
    \bibitem{hauser-amc}
    {Hauser} SC, {Headley} WC and {Michaels} AJ.
    \newblock {Signal detection effects on deep neural networks utilizing raw {IQ}
      for modulation classification}.
    \newblock In \emph{MILCOM 2017 - 2017 IEEE Military Communications Conference
      (MILCOM)}. pp. 121--127.
    \newblock \doi{10.1109/MILCOM.2017.8170853}.
    
    \bibitem{Hiremath_2018}
    {Hiremath} SM, {Deshmukh} S, {Rakesh} R et~al.
    \newblock {Blind Identification of Radio Access Techniques Based on
      Time-Frequency Analysis and Convolutional Neural Network}.
    \newblock In \emph{TENCON 2018 - 2018 IEEE Region 10 Conference}. pp.
      1163--1167.
    \newblock \doi{10.1109/TENCON.2018.8650355}.
    
    \bibitem{Tang_2018}
    {Tang} B, {Tu} Y, {Zhang} Z et~al.
    \newblock {Digital Signal Modulation Classification With Data Augmentation
      Using Generative Adversarial Nets in Cognitive Radio Networks}.
    \newblock \emph{IEEE Access} 2018; 6: 15713--15722.
    \newblock \doi{10.1109/ACCESS.2018.2815741}.
    
    \bibitem{adv_rfml:Davaslioglu2018a}
    {Davaslioglu} K and {Sagduyu} YE.
    \newblock {Generative Adversarial Learning for Spectrum Sensing}.
    \newblock In \emph{2018 IEEE International Conference on Communications (ICC)}.
      pp. 1--6.
    \newblock \doi{10.1109/ICC.2018.8422223}.
    
    \bibitem{Kulin_2018}
    {Kulin} M, {Kazaz} T, {Moerman} I et~al.
    \newblock {End-to-End Learning From Spectrum Data: A Deep Learning Approach for
      Wireless Signal Identification in Spectrum Monitoring Applications}.
    \newblock \emph{IEEE Access} 2018; 6: 18484--18501.
    \newblock \doi{10.1109/ACCESS.2018.2818794}.
    
    \bibitem{Wu18}
    {Wu} Y, {Li} X and {Fang} J.
    \newblock {A Deep Learning Approach for Modulation Recognition via Exploiting
      Temporal Correlations}.
    \newblock In \emph{2018 IEEE 19th International Workshop on Signal Processing
      Advances in Wireless Communications (SPAWC)}. pp. 1--5.
    \newblock \doi{10.1109/SPAWC.2018.8445938}.
    
    \bibitem{Li_2018}
    {Li} Z, {Liu} R, {Lin} X et~al.
    \newblock {Detection of Frequency-Hopping Signals Based on Deep Neural
      Networks}.
    \newblock In \emph{2018 IEEE 3rd International Conference on Communication and
      Information Systems (ICCIS)}. pp. 49--52.
    \newblock \doi{10.1109/ICOMIS.2018.8645029}.
    
    \bibitem{Tandiya_2018}
    {Tandiya} N, {Jauhar} A, {Marojevic} V et~al.
    \newblock {Deep Predictive Coding Neural Network for {RF} Anomaly Detection in
      Wireless Networks}.
    \newblock In \emph{2018 IEEE International Conference on Communications
      Workshops (ICC Workshops)}. pp. 1--6.
    \newblock \doi{10.1109/ICCW.2018.8403654}.
    
    \bibitem{Subekti_2018}
    {Subekti} A, {Pardede} HF, {Sustika} R et~al.
    \newblock {Spectrum Sensing for Cognitive Radio using Deep Autoencoder Neural
      Network and {SVM}}.
    \newblock In \emph{2018 International Conference on Radar, Antenna, Microwave,
      Electronics, and Telecommunications (ICRAMET)}. pp. 81--85.
    \newblock \doi{10.1109/ICRAMET.2018.8683930}.
    
    \bibitem{Jayaweera_2018}
    {Jayaweera} SK and {Aref} MA.
    \newblock {Cognitive Engine Design for Spectrum Situational Awareness and
      Signals Intelligence}.
    \newblock In \emph{2018 21st International Symposium on Wireless Personal
      Multimedia Communications (WPMC)}. pp. 478--483.
    \newblock \doi{10.1109/WPMC.2018.8712936}.
    
    \bibitem{oshea-2018-data}
    {O’Shea} TJ, {Roy} T and {Clancy} TC.
    \newblock {Over-the-Air Deep Learning Based Radio Signal Classification}.
    \newblock \emph{IEEE Journal of Selected Topics in Signal Processing} 2018;
      12(1): 168--179.
    
    \bibitem{Zhang_2018}
    {ZHANG} Y, {LIU} T, {ZHANG} L et~al.
    \newblock {A Deep Learning approach for Modulation Recognition}.
    \newblock In \emph{2018 IEEE 23rd International Conference on Digital Signal
      Processing (DSP)}. pp. 1--5.
    \newblock \doi{10.1109/ICDSP.2018.8631811}.
    
    \bibitem{Sang_2018}
    {Sang} Y and {Li} LA.
    \newblock {Application of novel architectures for Modulation Recognition}.
    \newblock In \emph{2018 IEEE Asia Pacific Conference on Circuits and Systems
      (APCCAS)}. pp. 159--162.
    \newblock \doi{10.1109/APCCAS.2018.8605691}.
    
    \bibitem{Vanhoy_2018}
    {Vanhoy} G, {Thurston} N, {Burger} A et~al.
    \newblock {Hierarchical Modulation Classification Using Deep Learning}.
    \newblock In \emph{MILCOM 2018 - 2018 IEEE Military Communications Conference
      (MILCOM)}. pp. 20--25.
    \newblock \doi{10.1109/MILCOM.2018.8599861}.
    
    \bibitem{Shapero_2018}
    {Shapero} SA, {Dill} AB and {Odelowo} BO.
    \newblock {Identifying Agile Waveforms with Neural Networks}.
    \newblock In \emph{2018 21st International Conference on Information Fusion
      (FUSION)}. pp. 745--752.
    \newblock \doi{10.23919/ICIF.2018.8455370}.
    
    \bibitem{Yashashwi_2019}
    {Yashashwi} K, {Sethi} A and {Chaporkar} P.
    \newblock {A Learnable Distortion Correction Module for Modulation
      Recognition}.
    \newblock \emph{IEEE Wireless Communications Letters} 2019; 8(1): 77--80.
    \newblock \doi{10.1109/LWC.2018.2855749}.
    
    \bibitem{Peng_2019}
    {Peng} S, {Jiang} H, {Wang} H et~al.
    \newblock {Modulation Classification Based on Signal Constellation Diagrams and
      Deep Learning}.
    \newblock \emph{IEEE Transactions on Neural Networks and Learning Systems}
      2019; 30(3): 718--727.
    \newblock \doi{10.1109/TNNLS.2018.2850703}.
    
    \bibitem{Wang_2019}
    {Wang} Y, {Liu} M, {Yang} J et~al.
    \newblock {Data-Driven Deep Learning for Automatic Modulation Recognition in
      Cognitive Radios}.
    \newblock \emph{IEEE Transactions on Vehicular Technology} 2019; 68(4):
      4074--4077.
    \newblock \doi{10.1109/TVT.2019.2900460}.
    
    \bibitem{Liu_2019}
    {Liu} H, {Zhu} X and {Fujii} T.
    \newblock {Cyclostationary based full-duplex spectrum sensing using adversarial
      training for convolutional neural networks}.
    \newblock In \emph{2019 International Conference on Artificial Intelligence in
      Information and Communication (ICAIIC)}. pp. 369--374.
    \newblock \doi{10.1109/ICAIIC.2019.8669026}.
    
    \bibitem{Zheng_2019}
    {Zheng} S, {Qi} P, {Chen} S et~al.
    \newblock {Fusion Methods for {CNN}-Based Automatic Modulation Classification}.
    \newblock \emph{IEEE Access} 2019; 7: 66496--66504.
    \newblock \doi{10.1109/ACCESS.2019.2918136}.
    
    \bibitem{Wang_ARL_2019}
    {Wang} P and {Vindiola} M.
    \newblock {Data Augmentation for Blind Signal Classification}.
    \newblock In \emph{MILCOM 2019 - 2019 IEEE Military Communications Conference
      (MILCOM)}. pp. 149--154.
    
    \bibitem{Clark19}
    {Clark} WH, {Arndorfer} V, {Tamir} B et~al.
    \newblock {Developing {RFML} Intuition: An Automatic Modulation Classification
      Architecture Case Study}.
    \newblock In \emph{MILCOM 2019 - 2019 IEEE Military Communications Conference
      (MILCOM)}. pp. 136--142.
    
    \bibitem{merchant_2019b}
    {Merchant} K and {Nousain} B.
    \newblock {Toward Receiver-Agnostic {RF} Fingerprint Verification}.
    \newblock In \emph{2019 IEEE Globecom Workshops (GC Wkshps)}. pp. 1--6.
    
    \bibitem{Moore_2020}
    {Moore} MO, {Clark IV} WH, {Beuhrer} RM et~al.
    \newblock {When is Enough Enough? {"}Just Enough{"} Decision Making with
      Recurrent Neural Networks for Radio Frequency Machine Learning}.
    \newblock In \emph{2020 IEEE 39th International Performance Computing and
      Communications Conference (IPCCC) (IEEE IPCCC 2020)}. Austin, USA.
    
    \bibitem{Cai_2010}
    {Cai} Q, {Chen} S, {Li} X et~al.
    \newblock {An integrated incremental self-organizing map and hierarchical
      neural network approach for cognitive radio learning}.
    \newblock In \emph{The 2010 International Joint Conference on Neural Networks
      (IJCNN)}. pp. 1--6.
    \newblock \doi{10.1109/IJCNN.2010.5596337}.
    
    \bibitem{ts_2014}
    {Torres-Sospedra} J, {Montoliu} R, {Martínez-Usó} A et~al.
    \newblock {{UJIIndoorLoc}: A new multi-building and multi-floor database for
      {WLAN} fingerprint-based indoor localization problems}.
    \newblock In \emph{2014 International Conference on Indoor Positioning and
      Indoor Navigation (IPIN)}. pp. 261--270.
    
    \bibitem{Kumar_2016}
    {Kumar} KAA.
    \newblock {{SoC} implementation of a modulation classification module for
      cognitive radios}.
    \newblock In \emph{2016 International Conference on Communication Systems and
      Networks (ComNet)}. pp. 87--92.
    \newblock \doi{10.1109/CSN.2016.7823992}.
    
    \bibitem{Schmidt_2017}
    {Schmidt} M, {Block} D and {Meier} U.
    \newblock {Wireless interference identification with convolutional neural
      networks}.
    \newblock In \emph{2017 IEEE 15th International Conference on Industrial
      Informatics (INDIN)}. pp. 180--185.
    \newblock \doi{10.1109/INDIN.2017.8104767}.
    
    \bibitem{Vyas_2017}
    {Vyas} MR, {Patel} DK and {Lopez-Benitez} M.
    \newblock {Artificial neural network based hybrid spectrum sensing scheme for
      cognitive radio}.
    \newblock In \emph{2017 IEEE 28th Annual International Symposium on Personal,
      Indoor, and Mobile Radio Communications (PIMRC)}. pp. 1--7.
    \newblock \doi{10.1109/PIMRC.2017.8292449}.
    
    \bibitem{Bitar_2017}
    {Bitar} N, {Muhammad} S and {Refai} HH.
    \newblock {Wireless technology identification using deep Convolutional Neural
      Networks}.
    \newblock In \emph{2017 IEEE 28th Annual International Symposium on Personal,
      Indoor, and Mobile Radio Communications (PIMRC)}. pp. 1--6.
    \newblock \doi{10.1109/PIMRC.2017.8292183}.
    
    \bibitem{oshea-detect1}
    {O'Shea} TJ, {Roy} T and {Erpek} T.
    \newblock {Spectral detection and localization of radio events with learned
      convolutional neural features}.
    \newblock In \emph{2017 25th European Signal Processing Conference (EUSIPCO)}.
      pp. 331--335.
    
    \bibitem{Fernandes_2018}
    {Fernandes} SS, {Makiuchi} MR, {Lamar} MV et~al.
    \newblock {An Adaptive Recurrent Neural Network Model Dedicated to
      Opportunistic Communication in Wireless Networks}.
    \newblock In \emph{2018 International Joint Conference on Neural Networks
      (IJCNN)}. pp. 01--08.
    \newblock \doi{10.1109/IJCNN.2018.8489720}.
    
    \bibitem{Testi_2018}
    {Testi} E, {Favarelli} E and {Giorgetti} A.
    \newblock {Machine Learning for User Traffic Classification in Wireless
      Systems}.
    \newblock In \emph{2018 26th European Signal Processing Conference (EUSIPCO)}.
      pp. 2040--2044.
    \newblock \doi{10.23919/EUSIPCO.2018.8553196}.
    
    \bibitem{Yi_2018}
    {Yi} S, {Wang} H, {Xue} W et~al.
    \newblock {Interference Source Identification for {IEEE} 802.15.4 wireless
      Sensor Networks Using Deep Learning}.
    \newblock In \emph{2018 IEEE 29th Annual International Symposium on Personal,
      Indoor and Mobile Radio Communications (PIMRC)}. pp. 1--7.
    \newblock \doi{10.1109/PIMRC.2018.8580857}.
    
    \bibitem{merchant_2018}
    {Merchant} K, {Revay} S, {Stantchev} G et~al.
    \newblock {Deep Learning for {RF} Device Fingerprinting in Cognitive
      Communication Networks}.
    \newblock \emph{IEEE Journal of Selected Topics in Signal Processing} 2018;
      12(1): 160--167.
    \newblock \doi{10.1109/JSTSP.2018.2796446}.
    
    \bibitem{Mohammed_2018}
    {Mohammed} S, {El Abdessamad} R, {Saadane} R et~al.
    \newblock {Performance Evaluation of Spectrum Sensing Implementation using
      Artificial Neural Networks and Energy Detection Method}.
    \newblock In \emph{2018 International Conference on Electronics, Control,
      Optimization and Computer Science (ICECOCS)}. pp. 1--6.
    \newblock \doi{10.1109/ICECOCS.2018.8610506}.
    
    \bibitem{Elbakly18}
    {Elbakly} R, {Aly} H and {Youssef} M.
    \newblock {{TrueStory}: Accurate and Robust {RF}-Based Floor Estimation for
      Challenging Indoor Environments}.
    \newblock \emph{IEEE Sensors Journal} 2018; 18(24): 10115--10124.
    
    \bibitem{Mendis_2019}
    {Mendis} GJ, {Wei} J and {Madanayake} A.
    \newblock {Deep Learning based Radio-Signal Identification with Hardware
      Design}.
    \newblock \emph{IEEE Transactions on Aerospace and Electronic Systems} 2019; :
      1--1\doi{10.1109/TAES.2019.2891155}.
    
    \bibitem{Hiremath_2019}
    {Hiremath} SM, {Behura} S, {Kedia} S et~al.
    \newblock {Deep Learning-Based Modulation Classification Using Time and
      {S}tockwell Domain Channeling}.
    \newblock In \emph{2019 National Conference on Communications (NCC)}. pp. 1--6.
    \newblock \doi{10.1109/NCC.2019.8732258}.
    
    \bibitem{AlHajri2019}
    {AlHajri} MI, {Ali} NT and {Shubair} RM.
    \newblock {Indoor Localization for {IoT} Using Adaptive Feature Selection: A
      Cascaded Machine Learning Approach}.
    \newblock \emph{IEEE Antennas and Wireless Propagation Letters} 2019; 18(11):
      2306--2310.
    
    \bibitem{Chawathe19}
    {Chawathe} SS.
    \newblock {Indoor-Location Classification Using {RF} Signatures}.
    \newblock In \emph{2019 IEEE 18th International Symposium on Network Computing
      and Applications (NCA)}. pp. 1--4.
    
    \bibitem{merchant_2019a}
    {Merchant} K and {Nousain} B.
    \newblock {Enhanced {RF} Fingerprinting for {IoT} Devices with Recurrent Neural
      Networks}.
    \newblock In \emph{MILCOM 2019 - 2019 IEEE Military Communications Conference
      (MILCOM)}. pp. 590--597.
    
    \bibitem{Simard}
    Simard PY, Steinkraus D and Platt JC.
    \newblock {Best practices for convolutional neural networks applied to visual
      document analysis}.
    \newblock In \emph{Seventh International Conference on Document Analysis and
      Recognition, 2003. Proceedings.} pp. 958--963.
    \newblock \doi{10.1109/ICDAR.2003.1227801}.
    
    \bibitem{NIPS20124824}
    Krizhevsky A, Sutskever I and Hinton GE.
    \newblock {ImageNet Classification with Deep Convolutional Neural Networks}.
    \newblock In Pereira F, Burges CJC, Bottou L et~al. (eds.) \emph{Advances in
      Neural Information Processing Systems 25}. Curran Associates, Inc., 2012.
    \newblock pp. 1097--1105.
    
    \bibitem{NIPS19961250}
    Yaeger LS, Lyon RF and Webb BJ.
    \newblock Effective training of a neural network character classifier for word
      recognition.
    \newblock In Mozer MC, Jordan MI and Petsche T (eds.) \emph{Advances in Neural
      Information Processing Systems 9}. MIT Press, 1997.
    \newblock pp. 807--816.
    
    \bibitem{rfml2019tutorial}
    {Flowers} B and {Headley} WC.
    \newblock {Adversarial Radio Frequency Machine Learning ({RFML}) with
      {PyTorch}, MILCOM 2019 - 2019 IEEE Military Communications Conference
      (MILCOM)}, 2019.
    
    \bibitem{adam_opt}
    Kingma DP and Ba JL.
    \newblock Adam: A method for stochastic optimization.
    \newblock San Diego, CA, United states.
    
    \bibitem{pytorch}
    Paszke A, Gross S, Massa F et~al.
    \newblock Pytorch: An imperative style, high-performance deep learning library.
    \newblock In Wallach H, Larochelle H, Beygelzimer A et~al. (eds.)
      \emph{Advances in Neural Information Processing Systems 32}. Curran
      Associates, Inc., 2019.
    \newblock pp. 8024--8035.
    \newblock
      \urlprefix\url{http://papers.neurips.cc/paper/9015-pytorch-an-imperative-style-high-performance-deep-learning-library.pdf}.
    
    \bibitem{numpy}
    Harris CR, Millman KJ, van~der Walt SJ et~al.
    \newblock Array programming with {NumPy}.
    \newblock \emph{Nature} 2020; 585(7825): 357--362.
    \newblock \doi{10.1038/s41586-020-2649-2}.
    \newblock \urlprefix\url{https://doi.org/10.1038/s41586-020-2649-2}.
    
    \bibitem{scipy}
    Virtanen P, Gommers R, Oliphant TE et~al.
    \newblock {{SciPy} 1.0: Fundamental Algorithms for Scientific Computing in
      Python}.
    \newblock \emph{Nature Methods} 2020; 17: 261--272.
    \newblock \doi{10.1038/s41592-019-0686-2}.
    
    \bibitem{swami_2000}
    Swami A and Sadler B.
    \newblock {Hierarchical digital modulation classification using cumulants}.
    \newblock \emph{Communications, IEEE Transactions on} 2000; 48(3): 416--429.
    \newblock \doi{10.1109/26.837045}.
    
    \bibitem{flowers_adv}
    {Flowers} B, {Buehrer} RM and {Headley} WC.
    \newblock {Evaluating Adversarial Evasion Attacks in the Context of Wireless
      Communications}.
    \newblock \emph{IEEE Transactions on Information Forensics and Security} 2020;
      15: 1102--1113.
    \newblock \doi{10.1109/TIFS.2019.2934069}.
    
    \bibitem{welch1947generalization}
    Welch BL.
    \newblock {The generalization of student's' problem when several different
      population variances are involved}.
    \newblock \emph{Biometrika} 1947; 34(1/2): 28--35.
    
    \bibitem{Sward}
    Sward W, Swanson T and Williams M.
    \newblock {Scintillation Simulator Test Results: Hardware-in-the-Loop Emulation
      of Ionospheric Scintillation}.
    \newblock In \emph{2014 IEEE Military Communications Conference}. pp.
      1361--1367.
    
\end{thebibliography}

\begin{biogs}
\textbf{William H. Clark IV} is a PhD candidate at Virginia Tech and a Research Associate with the Ted and Karyn Hume Center for National Security and Technology in Blacksburg, VA.\\

\noindent\textbf{Steven Hauser} is currently the Director of Engineering at Adapdix. He received his Bachelor of Science in Electrical Engineering from Virginia Tech in 2012, and his Master of Science in Electrical Engineering from Virginia Tech in 2018.\\

\noindent\textbf{Dr. William “Chris” Headley} is the Associate Director of the Electronic Systems Lab at the Virginia Tech Ted and Karyn Hume Center for National Security and Technology. He completed his PhD at Virginia Tech within the Bradley Department of Electrical and Computer Engineering in 2015. His research interests include radio frequency machine learning, spectrum sensing, and data generation.\\

\noindent\textbf{Alan J. Michaels} is a Research Professor and Director of the Virginia Tech Ted and Karyn Hume Center for National Security and Technology Electronic Systems Lab.  He earned his PhD from Georgia Tech and specializes in the development of secure communication systems.
\end{biogs}

\end{document}